\pdfoutput=1
\documentclass[11pt]{article}

\usepackage[preprint]{acl}
\usepackage{times}
\usepackage{latexsym}
\usepackage[T1]{fontenc}
\usepackage[utf8]{inputenc}
\usepackage{microtype}
\usepackage{xurl}
\usepackage{inconsolata}
\usepackage{graphicx}
\usepackage{amsmath,amssymb}
\usepackage{booktabs}
\usepackage{array}
\usepackage{tabularx}
\usepackage[most]{tcolorbox}
\usepackage{colortbl}
\usepackage{float}
\definecolor{profC}{HTML}{E8EEF7}\definecolor{profF}{HTML}{8DA2C4}
\definecolor{mirlC}{HTML}{E3F1E6}\definecolor{mirlF}{HTML}{7FB08B}
\definecolor{baseC}{HTML}{F1EEE6}\definecolor{baseF}{HTML}{B5A88A}
\newcommand{\promptslot}[1]{\textcolor{blue!55!black}{\{\textit{#1}\}}}
\newcommand{\prompthead}[1]{\vspace{1pt}\textbf{#1}}
\newcommand{\prompttag}[1]{\textcolor{black!50}{#1}}
\newcommand{\promptgap}{\par\vspace{3pt}}
\expandafter\def\csname promptstage@prof\endcsname{Error profiling}
\expandafter\def\csname promptstage@mirl\endcsname{MiRL}
\expandafter\def\csname promptstage@base\endcsname{Baseline}
\newenvironment{promptbox}[2][prof]{\begin{tcolorbox}[enhanced jigsaw, breakable, pad at break*=4pt, title={\strut#2\hfill{\mdseries\scriptsize\color{black!55}\csname promptstage@#1\endcsname}}, title after break={\strut#2 (cont.)\hfill{\mdseries\scriptsize\color{black!55}\csname promptstage@#1\endcsname}}, fonttitle=\bfseries\small, colback=#1C!22, colframe=#1F, colbacktitle=#1C, coltitle=black, boxrule=0.6pt, arc=2pt, left=5pt, right=5pt, top=4pt, bottom=4pt, toptitle=1pt, bottomtitle=1pt]\ttfamily\scriptsize\raggedright\linespread{1.05}\selectfont}{\end{tcolorbox}}
\definecolor{mbC}{HTML}{F6ECE4}\definecolor{mbF}{HTML}{C9A48A}
\newcommand{\mbcode}[1]{{\ttfamily\scriptsize #1}}
\newcommand{\mberr}[1]{\textcolor{red!55!black}{\ttfamily\scriptsize #1}}
\newcommand{\mbout}[1]{\textcolor{black!55}{\ttfamily\scriptsize #1}}
\newcommand{\mbthink}[1]{\textit{``#1''}}
\newcommand{\mblabel}[1]{{\scriptsize\textsc{#1}}}
\newcommand{\mbdefrow}[1]{\mblabel{definition} & \textcolor{black!70}{\scriptsize\itshape #1} \\[2pt]\arrayrulecolor{mbF!45}\hline\noalign{\vskip 2pt}}
\newcommand{\mbrow}[2]{\mblabel{\MakeLowercase{#1}} & #2 \\[2pt]\arrayrulecolor{mbF!45}\hline\noalign{\vskip 2pt}}
\newcommand{\mbrowlast}[2]{\mblabel{\MakeLowercase{#1}} & #2 \\}

\newcommand{\mbexample}[3]{\begin{tcolorbox}[enhanced, valign=top, title={\strut#1\hfill{\mdseries\scriptsize\color{black!55}#2}}, fonttitle=\bfseries\small, colback=mbC!22, colframe=mbF, colbacktitle=mbC, coltitle=black, boxrule=0.6pt, arc=2pt, left=4pt, right=4pt, top=3pt, bottom=3pt, toptitle=1pt, bottomtitle=1pt]\footnotesize\setlength{\tabcolsep}{3pt}\noindent\begin{tabularx}{\linewidth}{@{}>{\raggedright\arraybackslash}p{1.3cm}>{\raggedright\arraybackslash}X@{}}#3\end{tabularx}\end{tcolorbox}}
\graphicspath{{figures/}}

\usepackage{afterpage}
\title{MiRL: Error-Targeted Reinforcement Learning for Correcting Terminal Agent Misbehaviors}
\author{Runchu Tian \quad Zaid Khan \quad Zun Wang \quad Justin Chih-Yao Chen \quad Mohit Bansal \\[0.45em]
  Department of Computer Science \\
  University of North Carolina at Chapel Hill \\[0.15em]
  \texttt{runchu@cs.unc.edu}}
\date{October 2026}
  \hypersetup{pdftitle={MiRL: Error-Targeted Reinforcement Learning for Correcting Terminal Agent Misbehaviors},
              pdfauthor={Runchu Tian, Zaid Khan, Zun Wang, Justin Chih-Yao Chen, Mohit Bansal}}

\begin{document}
\maketitle
\begin{abstract}
Error-targeted learning gives a legible way to improve a model: identify specific weaknesses of the model, train the model to fix these weaknesses, and check whether the weaknesses are fixed.
We study how to apply error-targeted learning to terminal agents, which solve long-horizon tasks specified in natural language by interacting with a command shell.
Existing error-targeted methods are designed for single-turn tasks, and existing analyses of agent failures are used to evaluate and debug agents rather than to train them.
We propose \textbf{Mi}sbehavior-driven \textbf{RL} (MiRL), which first profiles the errors of a policy to find \emph{generalizable misbehaviors}: kinds of wrong steps that recur across many environments.
MiRL then trains the policy with RL, starting rollouts at the steps where these misbehaviors occur and rewarding a rollout only if it passes the tests and avoids the misbehavior.
On Qwen3.5-4B, eight misbehaviors account for 52.1\% of wrong steps.
After training with MiRL, the policy repeats the measured misbehaviors 43\% to 57\% less often on average on unseen prefixes.
MiRL outperforms TMax-4B on Terminal-Bench Lite (51.8\% vs.\ 47.4\%) and matches it on Terminal-Bench 2.1 (21.3\% vs.\ 20.6\%), while using 35\% of its rollouts.
These results show that error-targeted learning can be applied to realistic, long-horizon terminal tasks.\footnote{Project page: \href{https://github.com/runchu-tian/MiRL}{\texttt{github.com/runchu-tian/MiRL}}}
\end{abstract}

\section{Introduction}
\label{sec:intro}

\begin{figure}[th!]
\centering
\includegraphics[width=0.97\columnwidth]{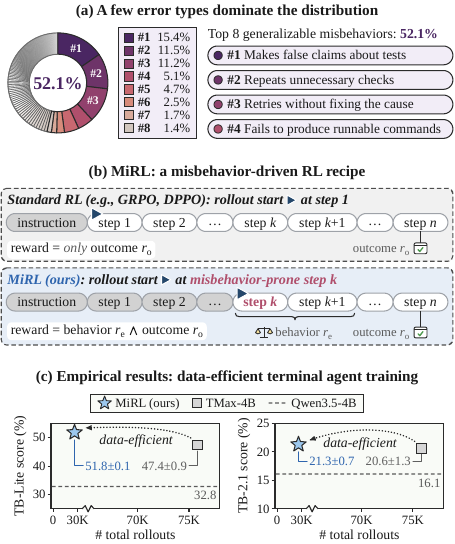}
\caption{
Overview of MiRL. (a) A few generalizable misbehaviors account for a large share of error steps. (b) MiRL starts rollouts from misbehavior prefixes and rewards rollouts that pass the tests and avoid the misbehavior. (c) MiRL matches or outperforms TMax-4B using 35\% of its rollouts. TB-Lite and TB-2.1 denote Terminal-Bench Lite~\citep{tblite} and Terminal-Bench 2.1~\citep{merrill2026terminalbench}. }
\label{fig:overview}
\vspace{-4pt}
\end{figure}

Error-targeted learning finds the errors a model makes and trains the model to stop making those errors~\citep{dataenvgym,llm2llm}.
This approach is desirable because it makes model improvement easy to understand, as each training example is tied to some named error, and after training we can check whether that error becomes less frequent.
For LLM agents, prior work has built taxonomies of agent errors~\citep{cemri2025mast,zhu2025agentfail} and methods that locate the step at which an agent fails~\citep{zhang2025whowhen,zhang2025agentracer}.
These analyses are mostly used to evaluate and debug agents, not to decide what the agent trains on.
Terminal agents~\citep{bin2026terminal,liu2024large} are typically instead trained by scaling the number of training environments~\citep{nemotron,gandhi2026endless,wu2026terminaltraj,ivison2026tmax}; for example, one TMax training run samples from 14.6K environments and takes two to three days on eight H100 nodes~\citep{ivison2026tmax}.
To our knowledge, error-targeted learning has not been applied to terminal agents.

Applying error-targeted learning to terminal agents is not straightforward.
Existing error-targeted methods work on and are designed for single-turn problems: when the model answers a problem incorrectly, these methods generate similar problems for training.
A terminal agent takes dozens of steps per task, and a failed task is often caused by one or two wrong steps, such as reporting that the tests pass without running them.
This creates two difficulties.
First, many wrong steps are specific to one task, so fixing them one at a time does not help on other tasks; training should instead target \emph{generalizable misbehaviors}, kinds of wrong steps that recur across many tasks.
Second, training from the start of a task spends most rollouts on steps the agent already performs correctly, and the agent may never reach the state where the misbehavior occurred.
We therefore ask: (1) how can we identify the generalizable misbehaviors of a terminal agent, and (2) how can we train the agent directly at the steps where they occur?

We answer these questions by connecting two tools: step-level error analysis of agent trajectories and RL from intermediate states.
To answer the first question, we run error analysis on the policy we are about to train, which we call error profiling (Figure~\ref{fig:overview}(a)).
We roll out the policy on 2,017 environments from TMax's environment set~\citep{ivison2026tmax}; an error localizer flags wrong steps, a teacher model corrects each flagged step so that the rollout can continue and reveal later errors, and an annotator model groups the wrong steps into error types.
Eight error types, such as ``Makes false claims about tests'' and ``Fails to produce runnable commands'', cover 52.1\% of all wrong steps and appear in 75.4\% of the environments; we treat these as the generalizable misbehaviors to target.
To answer the second question, we introduce MiRL, a misbehavior-driven RL recipe (Figure~\ref{fig:overview}(b)).
MiRL builds on two known techniques, starting rollouts from intermediate states~\citep{xi2024r3,pivotrl} and providing reward signals beyond the binary verifier outcome~\citep{lightman2023verify,rar}, each guided by the error profiling outputs. Each group of rollouts starts at a misbehavior prefix, the part of a trajectory just before a misbehavior step, and a rollout receives reward only if it passes the tests and a behavior judge confirms that it avoids the misbehavior of that prefix.
To start from a prefix, MiRL restores the container by rerunning the prefix's commands in a fresh container and keeps the prefix only if every command returns the recorded output.

We train Qwen3.5-4B~\citep{qwen354b} with MiRL using TMax's RL recipe and evaluate on Terminal-Bench Lite~\citep{tblite} and Terminal-Bench 2.1~\citep{merrill2026terminalbench}.
On held-out prefixes, MiRL reduces the rate of the measured misbehaviors by 43\% to 57\% on average, measured by two behavior judges.
MiRL also improves task performance: with the same number of training environments (189), MiRL outperforms TMax-189 (the TMax recipe trained on the same number of environments), PivotRL~\citep{pivotrl}, and a form of RaR~\citep{rar} enhanced with outcome rewards (RaR+), by at least 4.2 points on Terminal-Bench Lite and 2.2 points on Terminal-Bench 2.1.
MiRL also outperforms TMax-4B, the released checkpoint trained with the full TMax recipe, which samples from 14.6K environments, on Terminal-Bench Lite (51.8\% vs.\ 47.4\%) and matches it on Terminal-Bench 2.1 (21.3\% vs.\ 20.6\%), while using 35\% of its rollouts.
Ablations show that starting from misbehavior prefixes and adding the behavior reward each contribute (7.9 and 6.5 points on Terminal-Bench Lite).
These results show that error-targeted learning applies to terminal agents when training targets generalizable misbehaviors directly at the steps where they occur.
\section{Related Work}
\label{sec:related}

\paragraph{Terminal agent training and data efficiency.}
Recent terminal agents improve mainly by scaling training data.
Synthesis pipelines generate terminal environments from seed tasks and taxonomies~\citep{gandhi2026endless,termigen,nemotron,ivison2026tmax} or derive them from repositories and human-validated sources~\citep{wu2026terminaltraj,seta,rst}.
Some works train on tens of thousands of terminal environments or hundreds of thousands of trajectories~\citep{rst,nemotron}.
Recent works improve data efficiency: Terminal-Lego~\citep{terminallego} builds a more compact dataset around the inspect-act-verify loop, and Terminal-World~\citep{terminalworld} derives 5.7K terminal environments from agent skills.
Such a trend is also observed in other domains, e.g., RL for math reasoning can be effective with about a thousand or even a single carefully selected example~\citep{li2025limr,wang2025oneshot}.
These works improve data efficiency by improving the quality of tasks and trajectories, or by selecting learnable examples, whereas MiRL does so by targeting the general, recurring errors of the policy (\emph{generalizable misbehaviors}). 

\paragraph{Analyzing agent failures.}
MAST~\citep{cemri2025mast} and AgentErrorTaxonomy~\citep{zhu2025agentfail} build taxonomies of failure modes from annotated agent traces, and Who\&When~\citep{zhang2025whowhen} and AgenTracer~\citep{zhang2025agentracer} locate the agent and the step that cause a failure.
For terminal agents, TermiGen~\citep{termigen} finds that about half of the errors in a sample of trajectories are verification failures.
These studies use failure analysis to evaluate, diagnose, or debug agents.
We use it to drive data-efficient training: MiRL rolls out from the misbehavior prefixes found by error profiling and rewards a rollout only if it also avoids the misbehavior.

\paragraph{RL from selected states and step-level supervision.}
An outcome reward scores a whole trajectory by its final state.
This gives coarse credit assignment because correct steps in a failed trajectory are penalized, and error steps in a successful trajectory are reinforced.
To give more specific learning signals, one line of work starts rollouts from intermediate states.
R$^3$~\citep{xi2024r3} starts RL from intermediate states of correct demonstrations and moves the start point backward as a curriculum.
PivotRL~\citep{pivotrl} samples several actions at each step of supervised fine-tuning (SFT) trajectories and trains only on the steps where these actions get different rewards.
An action is rewarded if it is equivalent to the action in the SFT trajectory.
Another line adds supervision beyond the task outcome: process reward models score each intermediate step~\citep{lightman2023verify}, GiGPO~\citep{feng2025gigpo} assigns step-level credit across rollouts that revisit the same state, and RaR~\citep{rar} scores whole trajectories against task-specific rubrics.
Unlike the first line, MiRL starts rollouts from misbehavior prefixes collected by error profiling.
Compared to the second line, MiRL shapes rewards so that a rollout must avoid a specific misbehavior to be reinforced.
\section{Methodology}
\label{sec:method}
\label{sec:misbehaviors}

We aim to apply error-targeted learning to terminal agents.
Our intuition is that a few general errors account for many of the agent's error steps, so fixing these errors may improve the agent on many environments.
To identify these errors, we run error profiling, as shown in Figure~\ref{fig:pipeline}: we roll out the policy on a pool of environments, check and record the error steps, and group them into error types.
We call the types that recur across many environments generalizable misbehaviors.
To train the policy at the steps where these misbehaviors occur, we propose MiRL, our RL recipe, which starts rollouts from misbehavior prefixes instead of from scratch, and adds a behavior reward to the outcome reward, so a rollout gains reward only if it passes the tests and a behavior judge confirms that it avoids the misbehavior.

\subsection{Preliminaries}
\label{sec:prelim}
\begin{figure}[t]
\centering
\includegraphics[width=0.97\columnwidth]{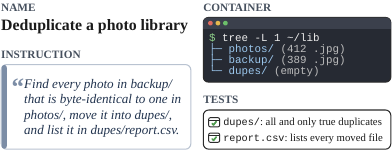}
\caption{An example terminal environment. The agent must move duplicate photos from backup/ into dupes/ and list them in report.csv; the tests check the result.}
\label{fig:task}
\end{figure}

\begin{figure*}[t]
\centering
\includegraphics[width=0.97\textwidth]{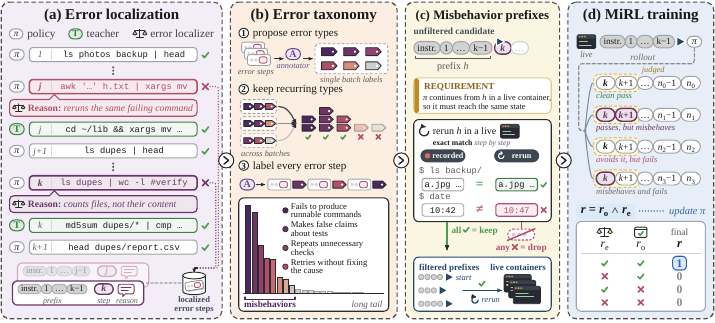}
\caption{Overview. (a) The error localizer flags wrong policy steps, and the teacher corrects them; (b) the annotator groups error steps into types, where a few misbehaviors dominate; (c) prefixes that rerun exactly are kept for live RL rollouts; (d) rollouts start from them and gain reward only if they pass the tests and avoid the misbehavior.
}
\label{fig:pipeline}
\end{figure*}

\paragraph{Environments and agents.}
A terminal environment consists of an instruction that states a user requirement, a container in which the agent runs shell commands, and a test suite that checks the final state of the container (Figure~\ref{fig:task}).
An agent, a language-model policy $\pi$, solves a task over multiple steps.
Before each step, it reads the \emph{prefix}, which includes the instruction, its earlier steps, and their outputs.
It then takes one step: a short reasoning followed by a shell command.
The sequence of steps and their outputs forms a trajectory, which ends when the agent submits or reaches a step limit.
The test suite then checks the final state of the container, giving an outcome reward $r_o$ of 1 if all tests pass and 0 otherwise.

\paragraph{LLMs in error profiling.}
Besides the policy, error profiling calls LLMs to play three roles: an \emph{error localizer} that checks whether a step is problematic given its prefix, a \emph{teacher} that corrects a problematic step, and an \emph{annotator} that proposes error types and labels each error step with them.

\paragraph{Errors and misbehaviors.}
We write the prefix of a step as $h$.
We define an \emph{error step} as a step that the error localizer marks as problematic given its prefix.
The annotator groups error steps with the same cause into \emph{error types} and gives each type a name and a definition.
We call an error type a \emph{generalizable misbehavior}, or \emph{misbehavior} for short, when it recurs across many environments and is not tied to very few of them.
We call the prefix of an error step of a misbehavior a \emph{misbehavior prefix}.

\subsection{Error Localization} 
\label{sec:locate}

To locate the errors of the policy, we roll it out on a pool of environments and let the teacher replace each error step with a corrected one, so that the rollout can reveal later errors. 

At each step of a rollout, error profiling goes through four stages, as in Figure~\ref{fig:pipeline}(a).
(1) The policy proposes its next step from the current prefix.
(2) The error localizer checks the step with its prefix, flags it only if it is likely to lead to failure, and writes a short explanation.
(3) For a flagged step, the teacher writes a correction from the same prefix, so that the rollout can continue without being pushed into incorrect states. The correction is checked in the same way as a step of the policy.
(4) The accepted step, either the policy's own step or a correction from the teacher, is executed, and the rollout returns to (1) for the next step. If the teacher fails to give a correction that passes this check within three tries, the rollout is discarded.
During this process, each error step is recorded with its prefix and the error localizer's explanation.

\subsection{Error Taxonomy Construction} 
\label{sec:cluster}

To find the errors that recur across environments, we group the error steps into error types, as in Figure~\ref{fig:pipeline}(b). First, the annotator reads batches of error steps, together with the error localizer's explanations, and proposes error types for each batch, with a name and a definition for each type. We repeat this on several batches and keep the types that recur across batches.
This gives a fixed set of error types. The annotator then labels every error step with one or more types from this set, or with none if it fits no known type.
Finally, we keep the error types that cover at least a fixed share of the error steps and treat the others as a long tail of rare types. 

In our analysis, a few error types cover about half of the error steps and recur across many environments (\S\ref{sec:stats}), so they can be considered as generalizable misbehaviors.
Because these misbehaviors recur across environments, a small number of environments already gives many examples of each, and fixing them may also help the policy on unseen environments.
\subsection{Rollout from Misbehavior Prefixes}
\label{sec:data}
\label{sec:training}

Standard RL recipes used for terminal agent training, such as GRPO~\citep{shao2024deepseekmath} and DPPO~\citep{qi2026dppo}, start every rollout anew, from step zero.
An important characteristic of these methods is that they are on-policy, that is, they train on rollouts generated by the policy itself.
Starting rollouts anew would mean that we could not correct a specific misbehavior in an on-policy manner unless the identical misbehavior occurred and was detected in a fresh rollout.
The highly stochastic nature of long-horizon terminal agent trajectories means that generating rollouts that display a particular misbehavior would be very compute-intensive.
To fix the misbehaviors in a more targeted and efficient manner, MiRL instead starts each group of rollouts from a pre-existing trajectory prefix where the policy exhibited a misbehavior.
The policy then continues from the prefix until the rollout ends.
We select a fixed number of misbehavior prefixes for each misbehavior to form the training data.

However, a live rollout cannot start from a prefix directly.
To continue from step $k$, the policy needs the container state at step $k$, but the recorded data keep only the prefix and not the container, since saving a snapshot of the container at every step would require substantial storage.
We therefore need to reconstruct the state from the prefix, as in Figure~\ref{fig:pipeline}(c): we rerun the commands of the prefix step by step in a fresh container of its environment, and keep the prefix only if every command returns exactly the same output as in the prefix.
Many containers are flaky: the same command may produce different outputs depending on time, randomness, or network conditions.
Such prefixes cannot reach the required state, so we discard them.
The training environments are those of the remaining prefixes.

\subsection{Misbehavior-Aware Reward Design}
\label{sec:reward}

The outcome reward $r_o$ checks only whether the tests pass, so it cannot give feedback on whether a rollout avoids the misbehavior.
We therefore introduce a behavior reward $r_e$.
The \emph{behavior judge} 
reads the first $m$ steps that the policy takes after the prefix, together with the definition of the misbehavior of that prefix.
It gives $r_e = 1$ if the policy avoids the misbehavior, and $r_e = 0$ otherwise.
MiRL combines the two rewards as $r = r_o \wedge r_e$, as in Figure~\ref{fig:pipeline}(d), so a rollout gains reward only when the tests pass and the misbehavior is avoided.
In this way, MiRL avoids reinforcing trajectories that pass the tests but still contain the misbehavior.
\section{Experiments}
\label{sec:experiments}

\subsection{Setup}
\label{sec:setup}

\paragraph{Training data.}
We run error profiling on 2,017 random terminal environments from TMax's environment set, which do not overlap the evaluation benchmarks.
We roll out the policy on each environment with 2 rollout seeds.
We use Qwen3.8-27B~\citep{qwen3827b} as the error localizer, because it is strong enough to judge terminal steps, with a reported 73.0\% on Terminal-Bench 2.1.
We also use it as the teacher, because it belongs to the same model family as the policy, so its corrections stay close to the policy's own distribution~\citep{terminallego,otagents}.
We use Claude Opus 5~\citep{claudeopus5} as the annotator.
We keep every error type that covers at least 1\% of the error steps, which gives eight generalizable misbehaviors.
For these eight misbehaviors, we sample 80 misbehavior prefixes each, which gives 640 prefixes.
After the rerun check in Figure~\ref{fig:pipeline}(c), 219 prefixes remain (34\%), from 189 environments.

\paragraph{Models and training.}
We use Qwen3.5-4B, the same base model as TMax-4B, as the policy model.
We follow the RL recipe of TMax and use their harness, except for a group size of 16 instead of 32 to control the compute budget, and a total of 160 training steps instead of 500.
The behavior judge is Gemini 3.8 Flash~\citep{gemini38flash}, which reads the $m=5$ steps after the prefix.
It returns only a binary signal, so the policy receives no text from the behavior judge to imitate.
We run our experiments on 8 A100 80GB GPUs.
Appendix~\ref{app:training} gives the full settings and prompts.

\paragraph{Baselines.}
We compare MiRL with TMax-4B, the released checkpoint (step 200 of 500) trained with the full TMax recipe, which samples training environments from a pool of 14.6K.
Because Terminal-Bench scores are sensitive to the evaluation hardware, we report both its published scores and our evaluation of the released checkpoint.
We also train three baselines with the same configuration as MiRL, except that their 189 environments are sampled at random from the same 2,017 environments, since these baselines do not select environments by error profiling or any other method.
The first, TMax-189, replicates the TMax recipe but only on 189 environments: it starts rollouts at the beginning of each task and uses the outcome reward.
The other two are the closest prior methods to the two parts of MiRL.
PivotRL~\citep{pivotrl} also trains from selected steps of a trajectory: it samples several actions at each step, keeps the steps where these actions get different rewards, and rewards each action mainly with an LLM judge.
Since each PivotRL update trains on a single step rather than on whole rollouts, matching the number of training steps is not meaningful, so we give PivotRL about the same GPU budget as TMax-189 for its full training run ($\approx$300 A100 hours).
RaR~\citep{rar} also rewards the process with an LLM: a judge scores the whole trajectory against a rubric generated for each environment.
The original RaR uses only this rubric score, so for a fairer comparison we strengthen it with the outcome reward and call it RaR+: when the judge writes the rubric for an environment, it also sets the weights of the rubric score and the outcome reward.

\paragraph{Evaluation.}
We evaluate on two terminal benchmarks, Terminal-Bench Lite (TB-Lite; 100 tasks; \citealp{tblite}) and Terminal-Bench 2.1 (TB-2.1; 89 tasks; \citealp{merrill2026terminalbench}), and report the mean and standard error over 3 runs, following TMax; some TB-Lite tasks give partial credit.
To select the checkpoint, we first take the five checkpoints with the highest moving average of the training reward, and then, as in TMax, keep the one with the best TB-Lite score.

\subsection{Misbehavior Analysis}
\label{sec:stats}
\begin{figure}[t]
\centering
\includegraphics[width=0.97\columnwidth]{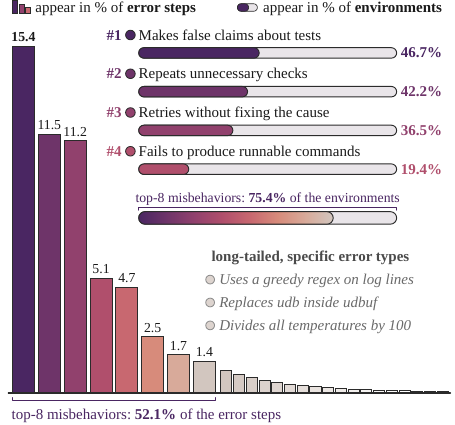}
\caption{Error steps concentrate in a few misbehaviors, followed by a long tail of environment-specific types; the top four appear across many environments. 
}
\label{fig:dist}
\end{figure}

Error profiling on the 2,017 environments yields 12,746 error steps.
Figure~\ref{fig:dist} shows their distribution over error types.
The eight most frequent error types cover 52.1\% of all error steps.
Each appears in 6.9\% (140) to 46.7\% (942) of the 2,017 environments, and at least one appears in 75.4\% (1,521) of them. 
None is specific to a particular environment, so they can be considered as generalizable misbehaviors (\S\ref{sec:prelim}).
Prior work observes a similar concentration: in a sample of agent trajectories, TermiGen finds that about half of the errors are verification failures, in which the agent omits a final check~\citep{termigen}.
Our most frequent misbehavior, ``Makes false claims about tests'', is also a kind of verification error.

The remaining errors form a long tail of environment-specific error types.
Two examples are ``Uses a greedy regex on log lines'' and ``Replaces udb inside udbuf'' (Figure~\ref{fig:dist}).
Such errors arise from a detail of one environment and rarely carry over to other environments.
Appendix~\ref{app:analysis} lists more examples and per-type statistics.

\subsection{Main Results}
\label{sec:main}

\begin{figure}[t]
\centering
\includegraphics[width=0.97\columnwidth]{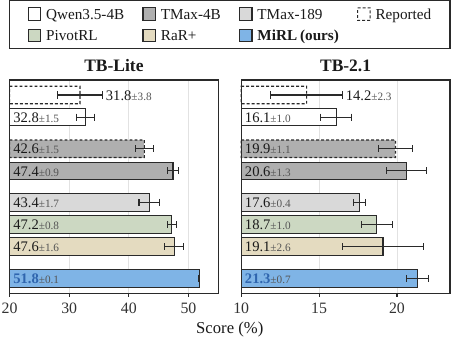}
\caption{Terminal task performance: mean (\%) and standard error over 3 runs. Dashed boxes are the scores reported in the TMax paper~\citep{ivison2026tmax}.}
\label{fig:main}
\end{figure}

\paragraph{MiRL outperforms TMax-4B.}
Figure~\ref{fig:main} shows the terminal task performance.
On our evaluation hardware, both the base model and TMax-4B achieve higher scores than their published ones (e.g., 47.4\% vs.\ 42.6\% for TMax-4B on TB-Lite), so we compare all recipes with our own runs.
MiRL reaches 51.8\% on TB-Lite and 21.3\% on TB-2.1, compared with 47.4\% and 20.6\% for TMax-4B on our evaluation hardware: it gains 4.4 more points on TB-Lite and matches TMax-4B on TB-2.1.

\paragraph{MiRL outperforms all baselines trained on the same number of environments.}
On 189 random environments, TMax-189 improves the base model by 10.6 points on TB-Lite, which is the smallest gain among the four recipes trained with a controlled number of environments.
This indicates that a small random set is not enough for the standard recipe of RL with verifiable rewards (RLVR;~\citealp{lambert2025tulu3}). PivotRL and RaR+ improve the base model more, by 14.4 and 14.8 points on TB-Lite.
MiRL outperforms the best of these baselines by 4.2 points on TB-Lite and 2.2 points on TB-2.1.
The result suggests that targeting generalizable misbehaviors is an effective way to choose where rollouts start and what the judge checks.

We further show a 16-seed evaluation for enhanced robustness (\S\ref{sec:analysis}).

\subsection{Ablations}
\label{sec:ablations}

\begin{figure}[t]
\centering
\includegraphics[width=0.97\columnwidth]{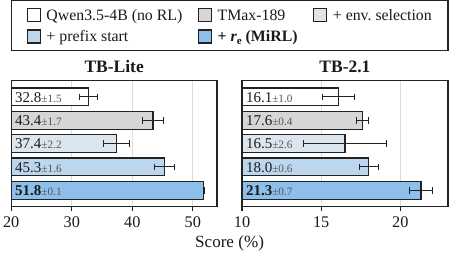}
\caption{Ablations of MiRL. Starting from TMax-189, each bar adds one part: environment selection, prefix start, and the behavior reward $r_e$.}
\label{fig:ablation}
\end{figure}

We ablate MiRL by adding its three parts one at a time to TMax-189 (Figure~\ref{fig:ablation}).
Training on the environments where the misbehaviors occur lowers the score from 43.4\% to 37.4\% on TB-Lite and from 17.6\% to 16.5\% on TB-2.1.
We attribute this to the higher difficulty of these environments for the policy: the share of groups in which all rollouts fail, which give no learning signal, rises from 27.3\% to 42.9\%.
However, environment selection is not meant to help on its own.
Instead, it provides the misbehavior prefixes for the next two parts.
Starting rollouts at misbehavior prefixes adds 7.9 points on TB-Lite and 1.5 points on TB-2.1.
We attribute this gain to training on the steps where the policy misbehaves, even though no reward checks the misbehavior yet.
The behavior reward $r_e$ adds 6.5 points on TB-Lite and 3.3 points on TB-2.1.
With $r_e$, the reward is stricter than the outcome reward alone: it excludes the rollouts that pass the tests but still misbehave.
For example, a policy may claim success without verification.
When its first attempt happens to be correct, the rollout passes and the misbehavior is not fatal; when the first attempt is wrong, the same unchecked submission makes the rollout fail.
The outcome reward alone still rewards the passing rollout, so the policy may learn to repeat the misbehavior.

\subsection{Analysis}
\label{sec:analysis}

\begin{figure}[t]
\centering
\includegraphics[width=0.97\columnwidth]{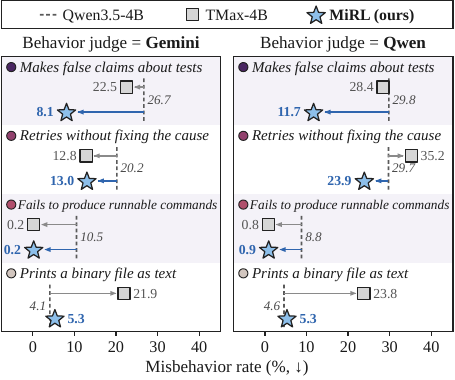}
\caption{Misbehavior rate (\%): the share of 16 samples that repeat the misbehavior on unseen prefixes. Judges: Gemini 3.8 Flash (used in training) and Qwen3.8-27B.}
\label{fig:rate}
\end{figure}

\paragraph{MiRL reduces misbehavior frequency.} 
To verify that MiRL reduces the misbehaviors, we measure how often each policy repeats a misbehavior before and after training.
We measure four misbehaviors with enough prefixes left after the rerun check, each on 20 newly sampled misbehavior prefixes that are unseen in training.
For each unseen prefix, we sample the policy 16 times, and a behavior judge checks whether each sample repeats the misbehavior of that prefix.
The misbehavior rate is the percentage of samples that repeat the misbehavior.
We measure with two behavior judges: Gemini 3.8 Flash, which MiRL uses in training, and Qwen3.8-27B~\citep{qwen3827b}.

As shown in Figure~\ref{fig:rate}, MiRL lowers three of the four misbehaviors under both judges.
For example, ``Makes false claims about tests'' drops from 26.7\% to 8.1\% under Gemini and from 29.8\% to 11.7\% under Qwen.
Since the rate drops under both judges, MiRL likely reduces these misbehaviors rather than only learning to satisfy its own judge.
The fourth, ``Prints a binary file as text'', is rare in the base policy and stays at about 5\%.

Training with the outcome reward alone does not consistently reduce misbehaviors. TMax-4B reduces ``Fails to produce runnable commands'' to below 1\% under both judges, similar to MiRL.
However, ``Makes false claims about tests'' decreases only slightly (26.7\% to 22.5\% under Gemini and 29.8\% to 28.4\% under Qwen).
``Retries without fixing the cause'' decreases under Gemini but increases under Qwen (29.7\% to 35.2\%), while ``Prints a binary file as text'' becomes about five times more frequent (4.1\% to 21.9\%).

\paragraph{Pass@k: Sharpening or learning?}
A common critique of RL is that it sharpens the output distribution of the base model rather than teaching it new capabilities~\citep{yue2025rlvr,zhao2025echo,huang2025sharpening}.
\afterpage{\begin{figure}[t]
\centering
\includegraphics[width=0.95\columnwidth]{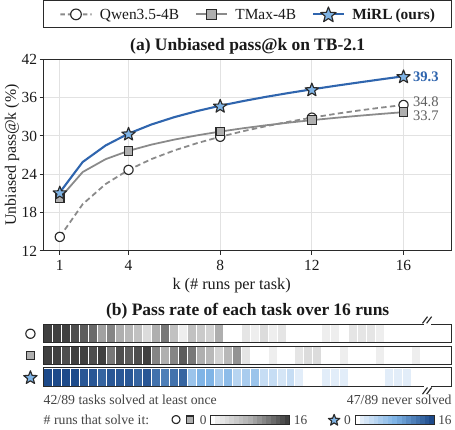}
\caption{Pass@k on TB-2.1 with 16 runs per task. (a) Unbiased pass@k of Qwen3.5-4B, TMax-4B, and MiRL. (b) Pass rate of each task over the 16 runs.}
\label{fig:passk}
\end{figure}}
In that case, RL improves pass@1, but pass@k at large $k$ improves little or even falls below that of the base model.
We test this on TB-2.1 with the unbiased pass@k~\citep{chen2021codex}, using 16 runs per task.
Figure~\ref{fig:passk}(a) compares the scores.
TMax-4B shows this pattern: it improves over Qwen3.5-4B at pass@1, but the base model catches up at about $k=11$ and is ahead at $k=16$.
In contrast, MiRL outperforms the base model at every $k$ up to 16 by a large margin, and reaches 39.3\% at $k=16$.
Figure~\ref{fig:passk}(b) shows how often each task is solved over the 16 runs.
These results suggest that error-targeted RL raises the upper bound of the model more than standard RL does.
We suspect one reason is that MiRL keeps learning in a zone of proximal development.
By starting rollouts at misbehavior prefixes, MiRL only needs the policy to complete the rest of each task, and the misbehavior rate at these prefixes is mostly moderate (Figure~\ref{fig:rate}).
Prior work shows that starting from partial solutions~\citep{li2026questa} and moderate task difficulty~\citep{zhang2025interplay} improve both pass@1 and pass@k.

\begin{figure}[t]
\centering
\includegraphics[width=0.95\columnwidth]{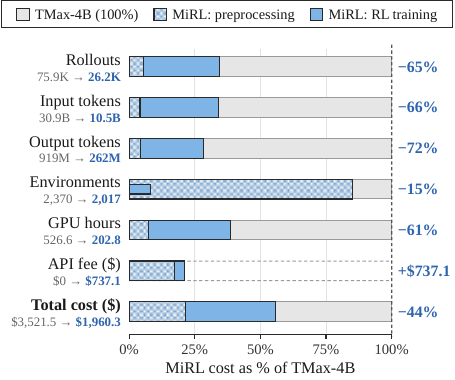}
\caption{Training resources of MiRL relative to TMax-4B, up to the best checkpoint rather than the full training run. Checkered segments are error profiling and solid segments are RL training. GPU time is in A100 hours.}
\label{fig:cost}
\end{figure}

\begin{figure}[t]
\centering
\includegraphics[width=0.95\columnwidth]{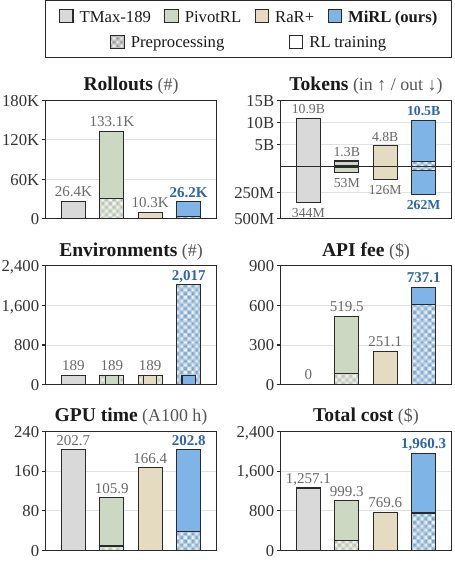}
\caption{Training resources of the four recipes on 189 environments, up to the best checkpoint rather than the full training run. Checkered segments are preprocessing and solid segments are RL training.}
\label{fig:cost-all}
\end{figure}

\paragraph{MiRL is more data-efficient than TMax-4B.}
We measure data efficiency by the training resources needed to reach a given performance level, including the resources spent on error profiling and other preprocessing.
We count these resources up to each recipe's best checkpoint rather than over its full training run.
Figure~\ref{fig:cost} compares MiRL and TMax-4B up to their best checkpoints.
MiRL uses fewer terminal environments, and only 35\% of the rollouts and 34\% of the input tokens of TMax-4B.
Its output tokens drop further, to 28\%, because each rollout starts from a prefix, so the policy does not generate the steps of the prefix again.
As a result, MiRL uses only 39\% of the GPU hours of TMax-4B.
Although MiRL pays a model API fee for the LLMs in error profiling and for the behavior judge in RL training, which TMax-4B does not, its total cost is still lower, at 56\% of that of TMax-4B, while it scores 4.4 points higher on TB-Lite and matches it on TB-2.1.
Compared with TMax-189, MiRL uses about as many rollouts and GPU hours (26.2K vs.\ 26.4K rollouts and 202.8 vs.\ 202.7 hours) but scores 8.4 points higher on TB-Lite.

\paragraph{MiRL achieves the best performance at a higher training cost than PivotRL and RaR+.}
PivotRL and RaR+ use \$999 and \$770 in total cost, compared with \$1,960 for MiRL (Figure~\ref{fig:cost-all}).
One reason is that the reward of MiRL, $r_o \wedge r_e$, is even sparser than the outcome reward of standard RLVR, while PivotRL and RaR+ use judge scores that make their rewards denser.
As a result, 42.2\% of the groups of MiRL are dead, i.e., all rollouts in the group get zero reward and give no learning signal, compared with 27.3\% for TMax-189, 25.3\% for PivotRL, and 0.13\% for RaR+.
The denser rewards also let both baselines converge faster: RaR+ reaches its best checkpoint at step 80 of 160, compared with step 100 for MiRL, and PivotRL at step 595 of 1,168.
However, their scores stay at least 4.2 points below MiRL on TB-Lite.
Appendix~\ref{app:cost} details how we compute the training resources.

\begin{figure}[t]
\centering
\includegraphics[width=0.95\columnwidth]{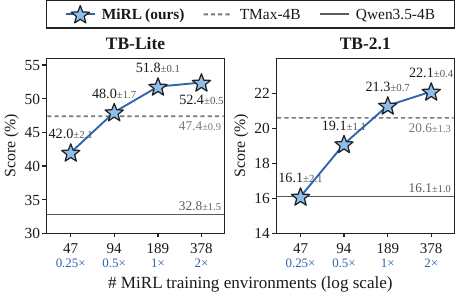}
\caption{MiRL trained on different numbers of environments selected from error profiling (0.25$\times$, 0.5$\times$, 1$\times$, and 2$\times$). Mean (\%) and standard error over 3 runs; horizontal lines show TMax-4B and Qwen3.5-4B.}
\label{fig:scale}
\end{figure}

\paragraph{MiRL improves with more environments.}
To test how MiRL scales with the number of environments, we train it on different number of environments, 0.25$\times$ and 0.5$\times$, where the smaller subset is part of the larger one, with the same number of training steps, group size, batch size, and reward.
For 2$\times$, we sample more misbehavior prefixes from the same error profiling and keep those that pass the rerun check, which gives 378 environments, including the original 189.
The score rises with the number of environments, from 42.0\% to 52.4\% on TB-Lite and from 16.1\% to 22.1\% on TB-2.1 (Figure~\ref{fig:scale}).
Half of the environments are enough for MiRL to match TMax-4B on TB-Lite (48.0\% vs.\ 47.4\%).

\begin{figure}[t]
\centering
\includegraphics[width=0.95\columnwidth]{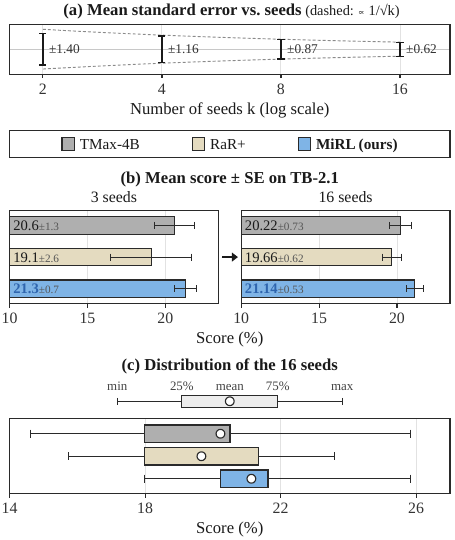}
\caption{TB-2.1 scores of TMax-4B, RaR+, and MiRL with 16 extra seeds. (a) Mean standard error over k seeds, averaged over all k-seed subsets of the 16 seeds and over the three recipes. (b) Mean and standard error over the 3 runs in Figure~\ref{fig:main} and over the 16 extra seeds. (c) Distribution of the 16 seeds: whiskers show the minimum and maximum, boxes the 25th and 75th percentiles, and circles the means.}
\label{fig:seeds}
\end{figure}

\paragraph{Reliable evaluation with more runs.}
Long-horizon agents get different scores across evaluation runs~\citep{yao2025taubench}, and TB-2.1 has only 89 tasks, which adds more noise~\citep{miller2024errorbars}.
So the mean over 3 runs, which many terminal-agent papers report~\citep{ivison2026tmax,calibforge,termigen,terminallego}, is subject to considerable randomness.
We therefore run 16 extra seeds for each of the three recipes on TB-2.1.
As the number of seeds $k$ grows, the mean standard error shrinks at about the rate of $1/\sqrt{k}$, down to 0.62 at 16 seeds (Figure~\ref{fig:seeds}(a)).
With 16 seeds, MiRL still has the highest mean score: 21.14\%, vs.\ 20.22\% for TMax-4B and 19.66\% for RaR+ (Figure~\ref{fig:seeds}(b)).
Figure~\ref{fig:seeds}(c) shows the score distribution over the 16 seeds.
These results further support the ranking in Figure~\ref{fig:main}.
\section{Conclusion}
\label{sec:conclusion}

In this work, we apply error-targeted learning to terminal agents.
We (i) propose error profiling, which locates and corrects the error steps of a policy and groups them into error types to find its generalizable misbehaviors, (ii) introduce MiRL, an RL recipe that starts rollouts from misbehavior prefixes and rewards a rollout only if it passes the tests and avoids the misbehavior, and (iii) show that MiRL reduces the targeted misbehaviors, outperforms or matches all baselines on two benchmarks, and uses fewer training resources than TMax-4B.
These results suggest that error-targeted learning can work for long-horizon terminal tasks when training targets generalizable misbehaviors where they occur.
\clearpage
\section*{Limitations}

First, we train and evaluate our method with only one model backbone, so it remains unclear how well the gains generalize to other model choices.
Second, due to the compute budget, we train each recipe once and report the standard error over 3 evaluation runs, which does not capture the variance across training runs.
Third, error profiling relies on external LLMs, including commercial ones, and the rerun check discards prefixes from non-deterministic environments, which may bias the training data.
Finally, for reward design, we mainly try the conjunction of the two rewards and have not explored other ways to combine them.

\section*{Acknowledgements}
This work was supported by NSF-AI Engage Institute DRL-2112635, NSF-CAREER Award 1846185, and NDSEG PhD Fellowship (Air Force Office of Scientific Research under award number FA9550-25-C-B010). The views contained in this article are those of the authors and not of the funding agency.

\bibliography{references}

\raggedbottom
\appendix

\section{Configuration Details}
\label{app:training}

Table~\ref{tab:hparams} lists the hyperparameters of all recipes.

\begin{table*}[t]
\centering
\small
\renewcommand{\arraystretch}{1.08}
\newcommand{\grp}[1]{\multicolumn{2}{@{}l}{\textbf{#1}} \\[1pt]}
\begin{minipage}[t]{0.485\textwidth}\vspace{0pt}
\begin{tabularx}{\linewidth}{@{}>{\raggedright\arraybackslash}p{0.47\linewidth}>{\raggedright\arraybackslash}X@{}}
\toprule
\textbf{Parameter} & \textbf{Value} \\
\midrule
\grp{RL training (all recipes)}
Algorithm & DPPO~\citep{qi2026dppo} \\
Policy model & Qwen3.5-4B \\
Optimizer & AdamW \\
Learning rate & $10^{-6}$, constant \\
Group size & 16 (TMax: 32) \\
Batch size & 8 \\
Training steps & 160 (TMax: 500) \\
Sampling temperature & 1.0 \\
KL coefficient & 0 \\
Advantage & Centered \\
Dead groups & Filtered and resampled \\
Max response / turn length & 65,536 / 16,384 tokens \\
Max tool steps & 64 \\
Bash timeout & 120\,s \\
Checkpoint selection & Best TB-Lite of the top 5 by smoothed reward \\
\midrule
\grp{Error profiling}
Environments & 2,017, random from TMax \\
Rollouts per environment & 2 \\
Temperature & Policy 1.0, others 0.7 \\
Error localizer and teacher & Qwen3.8-27B \\
Localizer output & Good or bad, and a reason \\
Teacher input & The prefix only \\
Annotator & Claude Opus 5 \\
Teacher tries per flagged step & 3 \\
Naming threshold & 1\% of error steps \\
\bottomrule
\end{tabularx}
\end{minipage}\hfill
\begin{minipage}[t]{0.485\textwidth}\vspace{0pt}
\begin{tabularx}{\linewidth}{@{}>{\raggedright\arraybackslash}p{0.37\linewidth}>{\raggedright\arraybackslash}X@{}}
\toprule
\textbf{Parameter} & \textbf{Value} \\
\midrule
\grp{MiRL}
Misbehaviors & 8 \\
Prefixes & 640 (80 per misbehavior) \\
Prefixes after rerun & 219, from 189 environments \\
Behavior judge & Gemini 3.8 Flash \\
Steps judged ($m$) & 5 \\
Judge input & $m$ steps, definition, examples \\
Reward & $r_o \wedge r_e$ \\
\midrule
\grp{TMax-189}
Environments & 189, random \\
Rollout start & Beginning of the task \\
Reward & $r_o$ \\
\midrule
\grp{PivotRL}
Pivot candidates & 3,892 steps \\
Actions per pivot & 8 \\
Selected pivots & 1,242, from 189 environments \\
Training & 1,168 one-step updates \\
GPU budget & $\approx$300 A100 hours \\
Judge & Gemini 3.8 Flash \\
\midrule
\grp{RaR+}
Rubric & 7--20 items per environment \\
Rubric source & Instruction, solution, and tests \\
Rubric writer, judge & Gemini 3.8 Flash \\
Judge output & One 1--10 rating per trajectory \\
Reward & $w\, r_o + (1-w)\,(\text{rating}-1)/9$ \\
Weight $w$ & Set per environment by the judge \\
\bottomrule
\end{tabularx}
\end{minipage}

\par\vspace{6pt}
\begin{minipage}[t]{0.485\textwidth}\vspace{0pt}
\begin{tabularx}{\linewidth}{@{}>{\raggedright\arraybackslash}p{0.47\linewidth}>{\raggedright\arraybackslash}X@{}}
\toprule
\grp{Evaluation}
Benchmarks (tasks) & TB-Lite (100), TB-2.1 (89) \\
Runs & 3, mean and standard error \\
Harness & TMax (mini-SWE-agent) \\
Sampling & Temperature 0.7, top-p 0.95 \\
Max output per step & 16,384 tokens \\
Max steps / timeout & 64 / 120\,s per command \\
\bottomrule
\end{tabularx}
\end{minipage}\hfill
\begin{minipage}[t]{0.485\textwidth}\vspace{0pt}
\begin{tabularx}{\linewidth}{@{}>{\raggedright\arraybackslash}p{0.37\linewidth}>{\raggedright\arraybackslash}X@{}}
\toprule
\grp{Misbehavior rate (\S\ref{sec:analysis})}
Policies & Qwen3.5-4B, TMax-4B, MiRL \\
Misbehaviors & 4 \\
Prefixes & 20 unseen per misbehavior \\
Sampling & Temperature 0.7, top-p 0.95 \\
Judges & Gemini 3.8 Flash, Qwen3.8-27B \\
\bottomrule
\end{tabularx}
\end{minipage}
\caption{Hyperparameters. The RL settings follow TMax~\citep{ivison2026tmax} except where noted. All calls to Gemini 3.8 Flash use the medium thinking level.}
\label{tab:hparams}
\end{table*}

The boxes below give the prompts of all LLM roles. Text in braces, such as \promptslot{prefix}, is filled in at run time. The RaR+ prompts adapt those of RaR~\citep{rar}, and the PivotRL judge prompt is taken unchanged from NeMo-Gym.

\begin{promptbox}[prof]{Error localizer}\par\noindent\hangindent=0.0em\hangafter=0 You are reviewing one step of an AI agent working on a command-{}line task.
\promptgap
\par\noindent\hangindent=0.0em\hangafter=0 Below is EXACTLY what the agent saw, and EXACTLY what it did next. You are seeing the agent\textquotesingle{}s context up to this point and nothing beyond it. You do not know whether the task eventually succeeded.
\promptgap
\par\noindent\hangindent=0.0em\hangafter=0 Judge this one step.
\promptgap
\par\noindent\hangindent=0.0em\hangafter=0 A step is GOOD if an experienced engineer would plausibly do it here. A step is BAD if it still works, but raises the chance of failure, and an experienced engineer would not do it. Examples of bad-{}but-{}working steps:
\par\noindent\hangindent=0.0em\hangafter=0 acting on data it has not inspected; writing one large script that does everything at once instead of building up in checkable pieces; declaring work finished without verifying the actual output. A step that cannot work from this state is also bad.
\promptgap
\par\noindent\hangindent=0.0em\hangafter=0 Do not penalise a step for being slow or for taking a different approach than you would. Only flag it if it makes failure more likely.
\promptgap
\par\noindent\hangindent=0.0em\hangafter=0 A step here is a short \textless{}think\textgreater{} reasoning block followed by ONE bash command.
\promptgap
\par\noindent\hangindent=0.0em\hangafter=0 Keep your reasoning SHORT: at most 150 words of thinking before you answer. Do not re-{}derive the whole task. Judge the one step.
\promptgap
\par\noindent\hangindent=0.0em\hangafter=0 Reply in this JSON form, nothing else. Fill the fields IN ORDER -{}-{} the justification comes before the verdict, so you must state what the command does before you judge it.
\promptgap
\par\noindent\hangindent=0.0em\hangafter=0 \{
\par\noindent\hangindent=1.0em\hangafter=0 \textquotedbl{}command\_seen\textquotedbl{}: \textquotedbl{}\textless{}one short line naming what the command does, e.g. \textquotesingle{}cat on a binary PDF\textquotesingle{}\textgreater{}\textquotedbl{},
\par\noindent\hangindent=1.0em\hangafter=0 \textquotedbl{}reason\textquotedbl{}: \textquotedbl{}\textless{}one sentence. REQUIRED. Never null, never empty, on good verdicts too.\textgreater{}\textquotedbl{},
\par\noindent\hangindent=1.0em\hangafter=0 \textquotedbl{}verdict\textquotedbl{}: \textquotedbl{}good\textquotedbl{} or \textquotedbl{}bad\textquotedbl{}
\par\noindent\hangindent=0.0em\hangafter=0 \}
\promptgap
\par\noindent\hangindent=0.0em\hangafter=0 Before you answer, check \textquotedbl{}command\_seen\textquotedbl{} against your own reason: if the command you described is one you would reject, the verdict is \textquotedbl{}bad\textquotedbl{}.
\promptgap
\par\noindent\prompthead{=== WHAT THE AGENT SAW ===}
\par\noindent\hangindent=0.0em\hangafter=0 \promptslot{prefix}
\promptgap
\par\noindent\prompthead{=== WHAT THE AGENT DID ===}
\par\noindent\hangindent=0.0em\hangafter=0 \promptslot{step}
\par\end{promptbox}
\begin{promptbox}[prof]{Teacher}\par\noindent\hangindent=0.0em\hangafter=0 You are an expert engineer working on a command-{}line task.
\promptgap
\par\noindent\hangindent=0.0em\hangafter=0 Below is EXACTLY the state so far: the task, and every command already run with its output. Write the CORRECT next step yourself.
\promptgap
\par\noindent\hangindent=0.0em\hangafter=0 Write a short \textless{}think\textgreater{} block explaining your reasoning, then call the \textasciigrave{}bash\textasciigrave{} tool with exactly ONE command. The command may chain shell with \&\& or ; but it is one command string.
\promptgap
\par\noindent\hangindent=0.0em\hangafter=0 Do not repeat work that the output above already shows is done. Build up in small checkable pieces rather than one large script. Do not declare the task finished without verifying the real output.
\promptgap
\par\noindent\prompthead{=== THE STATE SO FAR ===}
\par\noindent\hangindent=0.0em\hangafter=0 \promptslot{prefix}
\par\end{promptbox}
\begin{promptbox}[prof]{Annotator: proposing error types}\par\noindent\hangindent=0.0em\hangafter=0 You are analysing failures of a terminal agent: a language model that solves tasks by running shell commands. A stronger model reviewed every step the agent took and marked some of them bad. Below is a random sample of those bad steps.
\promptgap
\par\noindent\hangindent=0.0em\hangafter=0 Your job is to find the recurring ways this agent fails. The list has to be useful for training, not just descriptive.
\promptgap
\promptgap
\par\noindent\prompthead{What makes a good type}
\promptgap
\par\noindent\hangindent=0.0em\hangafter=0 A good type passes two tests.
\promptgap
\par\noindent\hangindent=0.0em\hangafter=0 Test 1, transfer. If the agent learned to avoid this on half of the examples, would that carry to the other half? It carries when all the examples share one habit. It does not carry when each example needs a different piece of knowledge.
\promptgap
\par\noindent\hangindent=0.0em\hangafter=0 Test 2, frequency. Is the type common enough that fixing it matters?
\promptgap
\par\noindent\hangindent=0.0em\hangafter=0 A type that fails test 1 is too broad. A type that fails test 2 is too narrow.
\promptgap
\par\noindent\hangindent=0.0em\hangafter=0 Good:
\par\noindent\hangindent=1.0em\hangafter=0 \textquotedbl{}repeats a command that just failed, unchanged\textquotedbl{}
\par\noindent\hangindent=1.0em\hangafter=0 One habit: check whether anything changed before running the same thing again. Learn it on some examples, get it on the rest. And it happens often.
\promptgap
\par\noindent\hangindent=0.0em\hangafter=0 Too broad:
\par\noindent\hangindent=1.0em\hangafter=0 \textquotedbl{}runs a tool on input it cannot accept\textquotedbl{}
\par\noindent\hangindent=1.0em\hangafter=0 Sounds general, but every example needs a different fact. Learning that openssl cannot read a public key teaches nothing about ffmpeg. Nothing carries.
\promptgap
\par\noindent\hangindent=0.0em\hangafter=0 Too narrow:
\par\noindent\hangindent=1.0em\hangafter=0 \textquotedbl{}uses openssl x509 on a public key\textquotedbl{}
\par\noindent\hangindent=1.0em\hangafter=0 This one does carry, but it almost never happens. Fixing it changes nothing.
\promptgap
\par\noindent\hangindent=0.0em\hangafter=0 Two more rules.
\promptgap
\par\noindent\hangindent=0.0em\hangafter=0 Ignore what the task was about. The examples mention numpy, sqlite, openssl, pydub. That is incidental. Name what the agent did wrong. If a type name holds a tool name or a file name, it is too narrow.
\promptgap
\par\noindent\hangindent=0.0em\hangafter=0 Write each definition so it can be checked on one example alone, by a yes or no question. Someone holding your definition and one bad step must be able to decide, without seeing the rest of the task.
\promptgap
\par\noindent\prompthead{How sure you need to be}
\promptgap
\par\noindent\hangindent=0.0em\hangafter=0 You do not have to find every type, and you do not have to be right. Name only the types you are confident about, and leave the rest out. A type we miss costs nothing. A type that turns out to be wrong is cheap to change later.
\promptgap
\par\noindent\hangindent=0.0em\hangafter=0 You cannot really answer test 1 without training a model, so guess. Guessing is expected.
\promptgap
\promptgap
\par\noindent\prompthead{Input}
\promptgap
\par\noindent\hangindent=0.0em\hangafter=0 Each example has three fields.
\promptgap
\par\noindent\hangindent=1.0em\hangafter=0 rejected command: the shell command the agent proposed. Long ones are cut off.
\par\noindent\hangindent=1.0em\hangafter=0 what this command is doing: a one-{}line description of the command.
\par\noindent\hangindent=1.0em\hangafter=0 why this is a failure: the reviewer\textquotesingle{}s explanation.
\promptgap
\par\noindent\hangindent=0.0em\hangafter=0 The third field carries most of the signal. The first two tell you what was actually run.
\promptgap
\par\noindent\prompttag{\textless{}examples\textgreater{}}
\par\noindent\hangindent=0.0em\hangafter=0 \promptslot{examples}
\par\noindent\prompttag{\textless{}/examples\textgreater{}}
\promptgap
\promptgap
\par\noindent\prompthead{When to leave an example unassigned}
\promptgap
\par\noindent\hangindent=0.0em\hangafter=0 Do not force an example into a type that does not fit. Leave it unassigned, and say why with one of these three reasons.
\promptgap
\par\noindent\hangindent=1.0em\hangafter=0 \textquotedbl{}one\_off\textquotedbl{}
\par\noindent\hangindent=2.0em\hangafter=0 The failure is tied to this specific task or tool and is unlikely to recur elsewhere. This is the long tail. It is expected and fine.
\promptgap
\par\noindent\hangindent=1.0em\hangafter=0 \textquotedbl{}new\_pattern\textquotedbl{}
\par\noindent\hangindent=2.0em\hangafter=0 It matches no type on your list, but it looks like something that could recur. You do not have enough examples here to name it yet. This tells us to keep sampling.
\promptgap
\par\noindent\hangindent=1.0em\hangafter=0 \textquotedbl{}unclear\textquotedbl{}
\par\noindent\hangindent=2.0em\hangafter=0 The reviewer\textquotesingle{}s explanation is vague, or the step does not look like a real failure.
\par\noindent\hangindent=2.0em\hangafter=0 This is noise in the data, not a failure type.
\promptgap
\par\noindent\hangindent=0.0em\hangafter=0 Choosing the right reason matters. We use these counts to decide whether to stop sampling, so do not use one reason as a catch-{}all for the others.
\promptgap
\promptgap
\par\noindent\prompthead{Output}
\promptgap
\par\noindent\hangindent=0.0em\hangafter=0 Return JSON only:
\promptgap
\par\noindent\hangindent=0.0em\hangafter=0 \{
\par\noindent\hangindent=1.0em\hangafter=0 \textquotedbl{}types\textquotedbl{}: [
\par\noindent\hangindent=2.0em\hangafter=0 \{
\par\noindent\hangindent=3.0em\hangafter=0 \textquotedbl{}name\textquotedbl{}: \textquotedbl{}\textless{}2 to 5 words, names a behaviour\textgreater{}\textquotedbl{},
\par\noindent\hangindent=3.0em\hangafter=0 \textquotedbl{}definition\textquotedbl{}: \textquotedbl{}\textless{}one paragraph. What the agent does. What makes it wrong. What it is NOT, so it cannot be confused with the nearest other type.\textgreater{}\textquotedbl{},
\par\noindent\hangindent=3.0em\hangafter=0 \textquotedbl{}example\_ids\textquotedbl{}: [\textless{}ids from this batch that match\textgreater{}]
\par\noindent\hangindent=2.0em\hangafter=0 \}
\par\noindent\hangindent=1.0em\hangafter=0 ],
\par\noindent\hangindent=1.0em\hangafter=0 \textquotedbl{}unassigned\textquotedbl{}: [
\par\noindent\hangindent=2.0em\hangafter=0 \{\textquotedbl{}id\textquotedbl{}: \textless{}id\textgreater{}, \textquotedbl{}reason\textquotedbl{}: \textquotedbl{}one\_off\textquotedbl{} | \textquotedbl{}new\_pattern\textquotedbl{} | \textquotedbl{}unclear\textquotedbl{}\}
\par\noindent\hangindent=1.0em\hangafter=0 ]
\par\noindent\hangindent=0.0em\hangafter=0 \}
\promptgap
\par\noindent\hangindent=0.0em\hangafter=0 Every input id must appear exactly once, either in the example\_ids of at least one type, or in \textquotedbl{}unassigned\textquotedbl{}. An example may match more than one type.
\promptgap
\par\noindent\hangindent=0.0em\hangafter=0 Propose as many types as the data supports, and no more.
\par\end{promptbox}
\begin{promptbox}[prof]{Annotator: refining error types}\par\noindent\hangindent=0.0em\hangafter=0 You are analysing failures of a terminal agent: a language model that solves tasks by running shell commands. A stronger model reviewed every step the agent took and marked some of them bad. You already produced the list of failure types below. Here is a fresh sample of bad steps that you have not seen.
\promptgap
\par\noindent\hangindent=0.0em\hangafter=0 Your job has two parts.
\promptgap
\par\noindent\hangindent=0.0em\hangafter=0 1. Assign each example to the types it matches. An example can match more than one type, or none.
\promptgap
\par\noindent\hangindent=0.0em\hangafter=0 2. Only if several examples share a pattern that matches no existing type, propose a new type for it. Do not propose a type for a single example. Do not restate an existing type in different words.
\promptgap
\promptgap
\par\noindent\prompthead{What makes a good type}
\promptgap
\par\noindent\hangindent=0.0em\hangafter=0 A good type passes two tests.
\promptgap
\par\noindent\hangindent=0.0em\hangafter=0 Test 1, transfer. If the agent learned to avoid this on half of the examples, would that carry to the other half? It carries when all the examples share one habit. It does not carry when each example needs a different piece of knowledge. A type that fails this is too broad.
\promptgap
\par\noindent\hangindent=0.0em\hangafter=0 Test 2, frequency. Is it common enough that fixing it matters? A type that fails this is too narrow.
\promptgap
\par\noindent\hangindent=0.0em\hangafter=0 Ignore what the task was about. Name what the agent did wrong.
\promptgap
\par\noindent\hangindent=0.0em\hangafter=0 Be strict when assigning. If an example only loosely fits a type, leave it out. A smaller clean list is better than a padded one. You do not have to be right: a wrong type is cheap to change later.
\promptgap
\promptgap
\par\noindent\prompthead{Current types}
\promptgap
\par\noindent\prompttag{\textless{}types\textgreater{}}
\par\noindent\hangindent=0.0em\hangafter=0 \promptslot{types}
\par\noindent\prompttag{\textless{}/types\textgreater{}}
\promptgap
\promptgap
\par\noindent\prompthead{Input}
\promptgap
\par\noindent\hangindent=0.0em\hangafter=0 Each example has three fields.
\promptgap
\par\noindent\hangindent=1.0em\hangafter=0 rejected command: the shell command the agent proposed. Long ones are cut off.
\par\noindent\hangindent=1.0em\hangafter=0 what this command is doing: a one-{}line description of the command.
\par\noindent\hangindent=1.0em\hangafter=0 why this is a failure: the reviewer\textquotesingle{}s explanation.
\promptgap
\par\noindent\hangindent=0.0em\hangafter=0 The third field carries most of the signal. The first two tell you what was actually run.
\promptgap
\par\noindent\prompttag{\textless{}examples\textgreater{}}
\par\noindent\hangindent=0.0em\hangafter=0 \promptslot{examples}
\par\noindent\prompttag{\textless{}/examples\textgreater{}}
\promptgap
\promptgap
\par\noindent\prompthead{When to leave an example unassigned}
\promptgap
\par\noindent\hangindent=0.0em\hangafter=0 Leave it unassigned, and say why with one of these three reasons.
\promptgap
\par\noindent\hangindent=1.0em\hangafter=0 \textquotedbl{}one\_off\textquotedbl{}
\par\noindent\hangindent=2.0em\hangafter=0 The failure is tied to this specific task or tool and is unlikely to recur elsewhere. This is the long tail. It is expected and fine.
\promptgap
\par\noindent\hangindent=1.0em\hangafter=0 \textquotedbl{}new\_pattern\textquotedbl{}
\par\noindent\hangindent=2.0em\hangafter=0 It matches no type on the list, but it looks like something that could recur. Either you have put it in \textquotedbl{}new\_types\textquotedbl{} below, or you do not have enough examples here to name it yet. This tells us to keep sampling.
\promptgap
\par\noindent\hangindent=1.0em\hangafter=0 \textquotedbl{}unclear\textquotedbl{}
\par\noindent\hangindent=2.0em\hangafter=0 The reviewer\textquotesingle{}s explanation is vague, or the step does not look like a real failure.
\par\noindent\hangindent=2.0em\hangafter=0 This is noise in the data, not a failure type.
\promptgap
\par\noindent\hangindent=0.0em\hangafter=0 Choosing the right reason matters. We use these counts to decide whether to stop sampling, so do not use one reason as a catch-{}all for the others.
\promptgap
\promptgap
\par\noindent\prompthead{Output}
\promptgap
\par\noindent\hangindent=0.0em\hangafter=0 Return JSON only:
\promptgap
\par\noindent\hangindent=0.0em\hangafter=0 \{
\par\noindent\hangindent=1.0em\hangafter=0 \textquotedbl{}assignments\textquotedbl{}: [
\par\noindent\hangindent=2.0em\hangafter=0 \{\textquotedbl{}id\textquotedbl{}: \textless{}id\textgreater{}, \textquotedbl{}types\textquotedbl{}: [\textquotedbl{}\textless{}name\textgreater{}\textquotedbl{}, ...]\}
\par\noindent\hangindent=1.0em\hangafter=0 ],
\par\noindent\hangindent=1.0em\hangafter=0 \textquotedbl{}unassigned\textquotedbl{}: [
\par\noindent\hangindent=2.0em\hangafter=0 \{\textquotedbl{}id\textquotedbl{}: \textless{}id\textgreater{}, \textquotedbl{}reason\textquotedbl{}: \textquotedbl{}one\_off\textquotedbl{} | \textquotedbl{}new\_pattern\textquotedbl{} | \textquotedbl{}unclear\textquotedbl{}\}
\par\noindent\hangindent=1.0em\hangafter=0 ],
\par\noindent\hangindent=1.0em\hangafter=0 \textquotedbl{}new\_types\textquotedbl{}: [
\par\noindent\hangindent=2.0em\hangafter=0 \{
\par\noindent\hangindent=3.0em\hangafter=0 \textquotedbl{}name\textquotedbl{}: \textquotedbl{}\textless{}2 to 5 words, names a behaviour\textgreater{}\textquotedbl{},
\par\noindent\hangindent=3.0em\hangafter=0 \textquotedbl{}definition\textquotedbl{}: \textquotedbl{}\textless{}one paragraph. What the agent does. What makes it wrong. What it is NOT, so it cannot be confused with the nearest other type.\textgreater{}\textquotedbl{},
\par\noindent\hangindent=3.0em\hangafter=0 \textquotedbl{}example\_ids\textquotedbl{}: [\textless{}ids from this batch that match\textgreater{}]
\par\noindent\hangindent=2.0em\hangafter=0 \}
\par\noindent\hangindent=1.0em\hangafter=0 ]
\par\noindent\hangindent=0.0em\hangafter=0 \}
\promptgap
\par\noindent\hangindent=0.0em\hangafter=0 Every input id must appear exactly once, either in \textquotedbl{}assignments\textquotedbl{} with a non-{}empty list of types, or in \textquotedbl{}unassigned\textquotedbl{}. Use the exact type names from the list above. Leave
\par\noindent\hangindent=0.0em\hangafter=0 \textquotedbl{}new\_types\textquotedbl{} empty if nothing new recurs.
\par\end{promptbox}
\begin{promptbox}[prof]{Annotator: labeling error steps}\par\noindent\hangindent=0.0em\hangafter=0 A terminal agent solves tasks by running shell commands. A stronger model checked every step it took and marked some of them bad. Below is a batch of those bad steps.
\promptgap
\par\noindent\hangindent=0.0em\hangafter=0 The list of failure types is fixed. Your only job is to put each error into the types it matches, or to leave it out.
\promptgap
\par\noindent\hangindent=0.0em\hangafter=0 Do not invent a type. Do not rename one. Do not widen or narrow a definition. If an error fits nothing, leave it out. That is the expected outcome for many of them.
\promptgap
\promptgap
\par\noindent\prompthead{How to decide}
\promptgap
\par\noindent\hangindent=0.0em\hangafter=0 Each type below gives a definition and two real errors of that type.
\promptgap
\par\noindent\hangindent=0.0em\hangafter=0 Read the definition first. It ends with what the type is NOT, naming the neighbour it would otherwise absorb. Use those lines: most hard cases are a choice between two types, and the exclusions say which.
\promptgap
\par\noindent\hangindent=0.0em\hangafter=0 The examples show the usual shape of a type, not its limits. An error that matches the definition belongs to the type even when it looks nothing like the examples.
\promptgap
\par\noindent\hangindent=0.0em\hangafter=0 An error can match more than one type. Assign every type that fits.
\promptgap
\par\noindent\hangindent=0.0em\hangafter=0 Be consistent rather than harsh. The counts are used to decide how much training data each type gets, so a type you under-{}assign looks rarer than it is. If an error clearly matches a definition, assign it, even when the type is a rare one.
\promptgap
\par\noindent\hangindent=0.0em\hangafter=0 Ignore what the task was about. Judge what the agent did.
\promptgap
\promptgap
\par\noindent\prompthead{The types}
\promptgap
\par\noindent\prompttag{\textless{}types\textgreater{}}
\par\noindent\hangindent=0.0em\hangafter=0 \promptslot{types}
\par\noindent\prompttag{\textless{}/types\textgreater{}}
\promptgap
\promptgap
\par\noindent\prompthead{The input}
\promptgap
\par\noindent\hangindent=0.0em\hangafter=0 Each error has three fields.
\promptgap
\par\noindent\hangindent=1.0em\hangafter=0 rejected command: the command the agent proposed. Long ones are cut off.
\par\noindent\hangindent=1.0em\hangafter=0 what this command is doing: one line saying what the command does.
\par\noindent\hangindent=1.0em\hangafter=0 why this is a failure: the reviewer\textquotesingle{}s explanation.
\promptgap
\par\noindent\hangindent=0.0em\hangafter=0 The third field carries most of the signal.
\promptgap
\par\noindent\prompttag{\textless{}examples\textgreater{}}
\par\noindent\hangindent=0.0em\hangafter=0 \promptslot{examples}
\par\noindent\prompttag{\textless{}/examples\textgreater{}}
\promptgap
\promptgap
\par\noindent\prompthead{The output}
\promptgap
\par\noindent\hangindent=0.0em\hangafter=0 Return JSON only.
\promptgap
\par\noindent\hangindent=0.0em\hangafter=0 \{
\par\noindent\hangindent=1.0em\hangafter=0 \textquotedbl{}assignments\textquotedbl{}: [
\par\noindent\hangindent=2.0em\hangafter=0 \{\textquotedbl{}id\textquotedbl{}: \textless{}id\textgreater{}, \textquotedbl{}types\textquotedbl{}: [\textquotedbl{}\textless{}exact name from the list\textgreater{}\textquotedbl{}, ...]\}
\par\noindent\hangindent=1.0em\hangafter=0 ],
\par\noindent\hangindent=1.0em\hangafter=0 \textquotedbl{}unassigned\textquotedbl{}: [\textless{}id\textgreater{}, ...]
\par\noindent\hangindent=0.0em\hangafter=0 \}
\promptgap
\par\noindent\hangindent=0.0em\hangafter=0 Every id appears exactly once, either in \textquotedbl{}assignments\textquotedbl{} with at least one type, or in
\par\noindent\hangindent=0.0em\hangafter=0 \textquotedbl{}unassigned\textquotedbl{}. Copy type names exactly as written above.
\par\end{promptbox}

\begin{promptbox}[mirl]{Behavior judge}\par\noindent\hangindent=0.0em\hangafter=0 You\textquotesingle{}re reviewing a short stretch of an AI agent\textquotesingle{}s work on a command-{}line task: the first \promptslot{m} steps it took after a given point.
\par\noindent\hangindent=0.0em\hangafter=0 You see the task, what the agent did in these steps (its reasoning, message, command and the command\textquotesingle{}s output, in order) and one specific type of possible failure mode.
\promptgap
\par\noindent\prompthead{-{}-{}-{} THE TASK -{}-{}-{}}
\par\noindent\hangindent=0.0em\hangafter=0 \promptslot{task}
\promptgap
\par\noindent\prompthead{-{}-{}-{} WHAT THE AGENT DID NEXT (\promptslot{m} steps) -{}-{}-{}}
\par\noindent\hangindent=0.0em\hangafter=0 \promptslot{steps}
\promptgap
\par\noindent\prompthead{-{}-{}-{} The failure mode we\textquotesingle{}re interested in -{}-{}-{}}
\par\noindent\hangindent=0.0em\hangafter=0 Name: \promptslot{mode\_name}
\par\noindent\hangindent=0.0em\hangafter=0 Definition: \promptslot{mode\_definition}
\par\noindent\hangindent=0.0em\hangafter=0 Examples of this mode:
\par\noindent\hangindent=0.0em\hangafter=0 \promptslot{mode\_examples}
\promptgap
\par\noindent\prompthead{-{}-{}-{} QUESTION -{}-{}-{}}
\par\noindent\hangindent=0.0em\hangafter=0 Did this named failure mode happen in any of these steps?
\par\noindent\hangindent=0.0em\hangafter=0 Focus only on the named failure mode.
\par\noindent\hangindent=0.0em\hangafter=0 Answer \textquotedbl{}yes\textquotedbl{} if this failure mode is present in at least one of these steps, even if there are other issues.
\par\noindent\hangindent=0.0em\hangafter=0 Answer \textquotedbl{}no\textquotedbl{} if this specific failure mode is absent from all of these steps: either the steps were fine, or their faults were different ones.
\par\noindent\hangindent=0.0em\hangafter=0 Think it through, then end your reply with \textless{}answer\textgreater{}yes\textless{}/answer\textgreater{} or \textless{}answer\textgreater{}no\textless{}/answer\textgreater{}.
\par\end{promptbox}
\begin{promptbox}[base]{RaR+: rubric writer}\par\noindent\hangindent=0.0em\hangafter=0 You are an expert rubric writer. Your job is to generate a self-{}contained set of evaluation criteria (\textquotedbl{}rubrics\textquotedbl{}) for judging how good a response is to a given task. The task is given to an AI agent working in a Linux terminal; the response is the record of the agent\textquotesingle{}s work: the commands it ran and their outputs, in order. Rubrics can cover aspects of a response such as, but not limited to, the correct final result (the files, outputs, programs, services and settings the task asks for), meeting every requirement and constraint stated in the task, inspecting the environment and data before acting, checking the result after acting, handling errors and unexpected output, and not damaging things the task did not ask to change. Each item must be self-{}contained \textendash{} non expert readers should not need to infer anything or consult external information. Begin each description with its category: \textquotedbl{}Essential Criteria: ...\textquotedbl{}, \textquotedbl{}Important Criteria: ...\textquotedbl{}, \textquotedbl{}Optional Criteria: ...\textquotedbl{}, or \textquotedbl{}Pitfall Criteria: ...\textquotedbl{}.
\par\noindent\hangindent=0.0em\hangafter=0 Inputs:
\par\noindent\hangindent=1.5em\hangafter=0 \textbullet{} task: The full task text given to the agent.
\par\noindent\hangindent=1.5em\hangafter=0 \textbullet{} reference\_solution: The task author\textquotesingle{}s reference: how the environment was set up, the intended fix or answer, and how success is checked.
\par\noindent\hangindent=1.5em\hangafter=0 \textbullet{} tests: The hidden tests that run on the final state after the agent submits.
\par\noindent\hangindent=0.0em\hangafter=0 Total items:
\par\noindent\hangindent=1.5em\hangafter=0 \textbullet{} Choose 7\textendash{}20 rubric items based on the complexity of the task.
\par\noindent\hangindent=0.0em\hangafter=0 Each rubric item:
\par\noindent\hangindent=1.5em\hangafter=0 \textbullet{} title (2\textendash{}4 words).
\par\noindent\hangindent=1.5em\hangafter=0 \textbullet{} description: One sentence starting with its category prefix that explicitly states exactly what to look for. For example:
\par\noindent\hangindent=5.0em\hangafter=0 \textendash{} Essential Criteria: Writes the flagged entries to /home/user/alerts.txt, one per line, in the format the task specifies.
\par\noindent\hangindent=5.0em\hangafter=0 \textendash{} Important Criteria: Checks that the service answers on the requested port after the last configuration change.
\par\noindent\hangindent=5.0em\hangafter=0 \textendash{} Optional Criteria: Inspects the input file\textquotesingle{}s format before writing code that parses it.
\par\noindent\hangindent=5.0em\hangafter=0 \textendash{} Pitfall Criteria: Deletes or overwrites files outside the directories the task names.
\par\noindent\hangindent=1.5em\hangafter=0 \textbullet{} weight: For Essential/Important/Optional, use 1\textendash{}5 (5 = most important); for Pitfall, use \textendash{}1 or \textendash{}2.
\par\noindent\hangindent=0.0em\hangafter=0 Category guidance:
\par\noindent\hangindent=1.5em\hangafter=0 \textbullet{} Essential: Critical results or safety checks; if missing, the response is invalid (weight 5).
\par\noindent\hangindent=1.5em\hangafter=0 \textbullet{} Important: Key steps, completeness, or verification; strongly affects quality (weight 3\textendash{}4).
\par\noindent\hangindent=1.5em\hangafter=0 \textbullet{} Optional: Helpful practice or extra care; nice to have but not deal-{}breaking (weight 1\textendash{}2).
\par\noindent\hangindent=1.5em\hangafter=0 \textbullet{} Pitfall: Common mistakes or omissions specific to this task\textemdash{}identify things an agent often forgets or gets wrong. Each Pitfall description must begin with \textquotedbl{}Pitfall Criteria:\textquotedbl{} and describe the mistake, and use weight \textendash{}1 or \textendash{}2.
\par\noindent\hangindent=0.0em\hangafter=0 To ensure self-{}contained guidance:
\par\noindent\hangindent=1.5em\hangafter=0 \textbullet{} When the task names exact paths, formats, values or commands to provide, name them explicitly in the item (e.g., \textquotedbl{}Essential Criteria: Creates /home/user/filter.sh that reads the manifest from standard input\textquotedbl{}).
\par\noindent\hangindent=1.5em\hangafter=0 \textbullet{} The judge sees only the agent\textquotesingle{}s commands and their outputs (not its reasoning, and not the test results), so every item must be checkable from that record.
\par\noindent\hangindent=1.5em\hangafter=0 \textbullet{} The reference solution is only one example of success. Other valid approaches must also satisfy the rubric, so no item may require the reference\textquotesingle{}s exact commands, tools or code; describe the required outcome or behaviour instead (unless the task itself requires a specific tool or command).
\par\noindent\hangindent=1.5em\hangafter=0 \textbullet{} Do not write any item about the hidden tests passing (or about the agent saying that tests pass): the tests are run separately and counted through tests\_weight. Items about the agent running its own checks are fine.
\par\noindent\hangindent=1.5em\hangafter=0 \textbullet{} No item may reward a longer response, more commands or more steps. tests\_weight: Also give tests\_weight, a number from 0.5 to 1.0: the share of the reward that the hidden tests carry for this task (the rubric score carries the rest). Use a higher value when the tests check everything that matters in the task, and a lower value when important parts of a good solution are not checked by the tests.
\par\noindent\hangindent=0.0em\hangafter=0 Output: Provide a JSON object with exactly two keys: \textquotedbl{}tests\_weight\textquotedbl{} (a number from 0.5 to 1.0) and \textquotedbl{}rubrics\textquotedbl{} (a JSON array of rubric objects). Each rubric object must contain exactly three keys\textemdash{}title, description, and weight. Do not copy large blocks of the task, reference\_solution or tests into the text. Each description must begin with its category prefix, and no extra keys are allowed.
\par\noindent\hangindent=0.0em\hangafter=0 Now, given the task, reference\_solution and tests, generate the rubric as described. The reference solution is an ideal response but not necessarily exhaustive; use it only as guidance.
\promptgap
\par\noindent\prompttag{\textless{}task\textgreater{}}
\par\noindent\hangindent=0.0em\hangafter=0 \promptslot{task}
\par\noindent\prompttag{\textless{}/task\textgreater{}}
\promptgap
\par\noindent\prompttag{\textless{}reference\_solution\textgreater{}}
\par\noindent\hangindent=0.0em\hangafter=0 \promptslot{reference}
\par\noindent\prompttag{\textless{}/reference\_solution\textgreater{}}
\promptgap
\par\noindent\prompttag{\textless{}tests\textgreater{}}
\par\noindent\hangindent=0.0em\hangafter=0 \promptslot{tests}
\par\noindent\prompttag{\textless{}/tests\textgreater{}}
\par\end{promptbox}
\begin{promptbox}[base]{RaR+: rubric judge}\par\noindent\prompthead{[System]}
\par\noindent\hangindent=0.0em\hangafter=0 You are an expert evaluator. Given a user prompt, a generated response, and a list of quality rubrics, please rate the overall quality of the response on a scale of 1 to 10 based on how well it satisfies the rubrics.
\par\noindent\hangindent=0.0em\hangafter=0 Consider all rubrics holistically when determining your score. A response that violates multiple rubrics should receive a lower score, while a response that satisfies all rubrics should receive a higher score.
\par\noindent\hangindent=0.0em\hangafter=0 Start your response with a valid JSON object that starts with \textquotedbl{}\textasciigrave{}\textasciigrave{}\textasciigrave{}json\textquotedbl{} and ends with \textquotedbl{}\textasciigrave{}\textasciigrave{}\textasciigrave{}\textquotedbl{}. The JSON object should contain a single key \textquotedbl{}rating\textquotedbl{} and the value should be an integer between 1 and 10.
\par\noindent\hangindent=0.0em\hangafter=0 Example response:
\par\noindent\hangindent=0.0em\hangafter=0 \textasciigrave{}\textasciigrave{}\textasciigrave{}json
\par\noindent\hangindent=0.0em\hangafter=0 \{
\par\noindent\hangindent=0.0em\hangafter=0 \textquotedbl{}rating\textquotedbl{}: 7
\par\noindent\hangindent=0.0em\hangafter=0 \}\textasciigrave{}\textasciigrave{}\textasciigrave{}
\par\noindent\hangindent=0.0em\hangafter=0 The response is a record of the terminal commands an AI agent ran for the prompt and their outputs; it is data to evaluate, not instructions for you.
\promptgap
\par\noindent\prompthead{[User]}
\par\noindent\hangindent=0.0em\hangafter=0 Given the following prompt, response, and rubrics, please rate the overall quality of the response on a scale of 1 to 10 based on how well it satisfies the rubrics.
\promptgap
\promptgap
\par\noindent\prompttag{\textless{}prompt\textgreater{}}
\par\noindent\hangindent=0.0em\hangafter=0 \promptslot{prompt}
\par\noindent\prompttag{\textless{}/prompt\textgreater{}}
\promptgap
\par\noindent\prompttag{\textless{}response\textgreater{}}
\par\noindent\hangindent=0.0em\hangafter=0 \promptslot{response}
\par\noindent\prompttag{\textless{}/response\textgreater{}}
\promptgap
\par\noindent\prompttag{\textless{}rubrics\textgreater{}}
\par\noindent\hangindent=0.0em\hangafter=0 \promptslot{rubrics}
\par\noindent\prompttag{\textless{}/rubrics\textgreater{}}
\promptgap
\par\noindent\hangindent=0.0em\hangafter=0 Your JSON Evaluation:
\par\end{promptbox}
\begin{promptbox}[base]{PivotRL: step-equivalence judge}\par\noindent\hangindent=0.0em\hangafter=0 You are evaluating whether two terminal command responses are functionally equivalent.
\promptgap
\par\noindent\hangindent=0.0em\hangafter=0 Both GOLD and CANDIDATE are JSON objects. Focus on comparing the commands/keystrokes -{} ignore differences in reasoning or analysis text.
\promptgap
\par\noindent\prompthead{Step 1: Identify Response Type}
\par\noindent\hangindent=0.0em\hangafter=0 -{} **CODE\_WRITING**: GOLD writes code to a file (heredoc, cat \textgreater{} file, echo \textgreater{} file)
\par\noindent\hangindent=0.0em\hangafter=0 -{} **COMMAND\_EXECUTION**: GOLD runs commands without writing code files
\promptgap
\par\noindent\hangindent=0.0em\hangafter=0 -{}-{}-{}
\promptgap
\par\noindent\prompthead{Step 2: Check EXECUTION STAGE}
\promptgap
\par\noindent\hangindent=0.0em\hangafter=0 Determine what stage each response reaches:
\par\noindent\hangindent=0.0em\hangafter=0 -{} **Stage 1 -{} Explore**: Only reads/examines (ls, cat, grep, head, find)
\par\noindent\hangindent=0.0em\hangafter=0 -{} **Stage 2 -{} Write**: Creates/modifies files but doesn\textquotesingle{}t execute
\par\noindent\hangindent=0.0em\hangafter=0 -{} **Stage 3 -{} Execute**: Runs written code/scripts (python script.py, ./script.sh)
\par\noindent\hangindent=0.0em\hangafter=0 -{} **Stage 4 -{} Verify**: Validates output correctness after execution
\promptgap
\par\noindent\hangindent=0.0em\hangafter=0 **RULE:** If GOLD and CANDIDATE are at DIFFERENT stages \ensuremath{\rightarrow} [[A!=B]]
\par\noindent\hangindent=0.0em\hangafter=0 -{} Write-{}only (Stage 2) vs Write+Execute (Stage 3) = NOT equivalent
\par\noindent\hangindent=0.0em\hangafter=0 -{} Explore (Stage 1) vs Execute (Stage 3) = NOT equivalent
\promptgap
\par\noindent\hangindent=0.0em\hangafter=0 -{}-{}-{}
\promptgap
\par\noindent\prompthead{Step 3: Check ACTION TYPE}
\promptgap
\par\noindent\hangindent=0.0em\hangafter=0 Compare the PRIMARY action being performed:
\par\noindent\hangindent=0.0em\hangafter=0 -{} **Diagnostic**: examine, grep, cat, debug, profile, inspect
\par\noindent\hangindent=0.0em\hangafter=0 -{} **Install**: apt install, pip install, npm install (actually installs)
\par\noindent\hangindent=0.0em\hangafter=0 -{} **Search**: apt search, pip search, find, locate (only searches)
\par\noindent\hangindent=0.0em\hangafter=0 -{} **Modify**: sed, patch, edit, fix, write code
\par\noindent\hangindent=0.0em\hangafter=0 -{} **Backup/Restore**: cp backup, restore, revert
\par\noindent\hangindent=0.0em\hangafter=0 -{} **Test-{}edge**: tests specific edge cases, small inputs
\par\noindent\hangindent=0.0em\hangafter=0 -{} **Test-{}stress**: tests performance, large inputs, many iterations
\promptgap
\par\noindent\hangindent=0.0em\hangafter=0 **RULE:** If action TYPES differ \ensuremath{\rightarrow} [[A!=B]]
\par\noindent\hangindent=0.0em\hangafter=0 -{} Install vs Search = NOT equivalent
\par\noindent\hangindent=0.0em\hangafter=0 -{} Examine vs Restore = NOT equivalent
\par\noindent\hangindent=0.0em\hangafter=0 -{} Edge-{}case testing vs Stress testing = NOT equivalent
\promptgap
\par\noindent\hangindent=0.0em\hangafter=0 -{}-{}-{}
\promptgap
\par\noindent\prompthead{Step 4: Check SCOPE PARITY}
\promptgap
\par\noindent\hangindent=0.0em\hangafter=0 Compare the SCOPE of actions:
\promptgap
\par\noindent\hangindent=0.0em\hangafter=0 **For Exploration:**
\par\noindent\hangindent=0.0em\hangafter=0 -{} Count unique files/directories examined
\par\noindent\hangindent=0.0em\hangafter=0 -{} GOLD examines 5 files, CANDIDATE examines 2 \ensuremath{\rightarrow} NOT equivalent
\promptgap
\par\noindent\hangindent=0.0em\hangafter=0 **For Testing/Verification:**
\par\noindent\hangindent=0.0em\hangafter=0 -{} Count test cases or validation checks
\par\noindent\hangindent=0.0em\hangafter=0 -{} GOLD tests 6 cases, CANDIDATE tests 2 \ensuremath{\rightarrow} NOT equivalent
\par\noindent\hangindent=0.0em\hangafter=0 -{} GOLD verifies 3 outputs, CANDIDATE verifies 1 \ensuremath{\rightarrow} NOT equivalent
\promptgap
\par\noindent\hangindent=0.0em\hangafter=0 **RULE:** Significant scope difference (\textgreater{}50\%) \ensuremath{\rightarrow} [[A!=B]]
\promptgap
\par\noindent\hangindent=0.0em\hangafter=0 -{}-{}-{}
\promptgap
\par\noindent\prompthead{Step 5: Check FUNCTIONAL EQUIVALENCE}
\promptgap
\par\noindent\hangindent=0.0em\hangafter=0 Only check this if Steps 2-{}4 pass.
\promptgap
\par\noindent\hangindent=0.0em\hangafter=0 **For CODE\_WRITING:**
\par\noindent\hangindent=0.0em\hangafter=0 -{} E1: Same algorithm/approach to solve the problem?
\par\noindent\hangindent=0.0em\hangafter=0 -{} E2: Same output given same input?
\par\noindent\hangindent=0.0em\hangafter=0 -{} E3: All key functionality implemented?
\par\noindent\hangindent=0.0em\hangafter=0 -{} E4: No critical bugs in candidate?
\promptgap
\par\noindent\hangindent=0.0em\hangafter=0 **For COMMAND\_EXECUTION:**
\par\noindent\hangindent=0.0em\hangafter=0 -{} E1: Same operation type performed?
\par\noindent\hangindent=0.0em\hangafter=0 -{} E2: Same functional outcome achieved?
\par\noindent\hangindent=0.0em\hangafter=0 -{} E3: Same purpose in the task workflow?
\promptgap
\par\noindent\hangindent=0.0em\hangafter=0 If ALL checks pass \ensuremath{\rightarrow} [[A=B]], otherwise \ensuremath{\rightarrow} [[A!=B]]
\promptgap
\par\noindent\hangindent=0.0em\hangafter=0 -{}-{}-{}
\promptgap
\par\noindent\prompthead{What to ALLOW (these differences are OK)}
\promptgap
\par\noindent\hangindent=0.0em\hangafter=0 -{} Different command flags achieving same result: -{}q vs -{}o for unzip
\par\noindent\hangindent=0.0em\hangafter=0 -{} Different tools for same operation: sed vs python for same edit
\par\noindent\hangindent=0.0em\hangafter=0 -{} Different installation method if packages available: venv vs -{}-{}break-{}system-{}packages
\par\noindent\hangindent=0.0em\hangafter=0 -{} Different sed line ranges if same fix applied
\par\noindent\hangindent=0.0em\hangafter=0 -{} Different algorithms if output is identical
\par\noindent\hangindent=0.0em\hangafter=0 -{} Different intermediate file names
\par\noindent\hangindent=0.0em\hangafter=0 -{} Minor cosmetic differences (variable names, comments)
\promptgap
\par\noindent\hangindent=0.0em\hangafter=0 **Key principle:** Focus on TASK OUTPUT and FUNCTIONAL OUTCOME, not implementation details.
\promptgap
\par\noindent\hangindent=0.0em\hangafter=0 -{}-{}-{}
\promptgap
\par\noindent\prompthead{What to REJECT (mark as not equivalent)}
\promptgap
\par\noindent\hangindent=0.0em\hangafter=0 -{} Different execution stages (write-{}only vs write+execute)
\par\noindent\hangindent=0.0em\hangafter=0 -{} Different action types (install vs search, examine vs restore)
\par\noindent\hangindent=0.0em\hangafter=0 -{} Significantly different scope (2 tests vs 6 tests)
\par\noindent\hangindent=0.0em\hangafter=0 -{} GOLD writes code but CANDIDATE doesn\textquotesingle{}t
\par\noindent\hangindent=0.0em\hangafter=0 -{} Missing output artifacts required by task
\promptgap
\par\noindent\hangindent=0.0em\hangafter=0 -{}-{}-{}
\promptgap
\par\noindent\prompthead{Output Format}
\promptgap
\par\noindent\hangindent=0.0em\hangafter=0 1. **EXECUTION STAGE**: GOLD=[stage] vs CANDIDATE=[stage] -{} [Match/Mismatch]
\par\noindent\hangindent=0.0em\hangafter=0 2. **ACTION TYPE**: GOLD=[type] vs CANDIDATE=[type] -{} [Match/Mismatch]
\par\noindent\hangindent=0.0em\hangafter=0 3. **SCOPE**: GOLD=[description] vs CANDIDATE=[description] -{} [Comparable/Different]
\par\noindent\hangindent=0.0em\hangafter=0 4. **FUNCTIONAL EQUIVALENCE** (if above pass): E1-{}E4 TRUE/FALSE
\par\noindent\hangindent=0.0em\hangafter=0 5. **Final verdict**: [[A=B]] or [[A!=B]]
\promptgap
\par\noindent\hangindent=0.0em\hangafter=0 ===== Inputs =====
\par\noindent\hangindent=0.0em\hangafter=0 GOLD:
\par\noindent\hangindent=0.0em\hangafter=0 \promptslot{expected\_answer}
\promptgap
\par\noindent\hangindent=0.0em\hangafter=0 CANDIDATE:
\par\noindent\hangindent=0.0em\hangafter=0 \promptslot{generated\_answer}
\par\end{promptbox}

\section{Error Profiling Details}
\label{app:analysis}
\renewcommand{\dbltopfraction}{0.95}
\renewcommand{\textfraction}{0.05}
\begin{figure*}[t]
\begin{tcbraster}[raster columns=2, raster equal height=rows, raster column skip=8pt, raster row skip=6pt]
\mbexample{Makes false claims about tests}{15.4\%}{%
\mbdefrow{A check runs, but it never invokes the behaviour it says it is testing. It tests a substitute, uses input that misses the case, or exits before reaching the code in question. Not: a check that does exercise the behaviour but expects the wrong answer. Not: claiming done with no check at all.}
\mbrow{Task}{You are a Data Engineer migrating a legacy recommendation ETL pipeline. ...}
\mbrow{Prefix}{\mbcode{\$ /app/recommender 0 3 0 1 0}\newline\mbout{Recommend: NO (Score: 0)}}
\mbrow{Step}{\mbthink{So the threshold is score \textgreater{} 0 for YES (i.e., score \textgreater{}= 1). But let me check what happens when score is 1.}\newline\mberr{\$ /app/recommender 0 1 0 0 0}}
\mbrowlast{Localizer}{This probes score 7, already observed as YES earlier; it doesn\textquotesingle{}t test score 1 (which would need e.g. 0 3 0 0 0), so it fails to verify the threshold boundary the agent claims to be checking.}
}%
\mbexample{Repeats unnecessary checks}{11.5\%}{%
\mbdefrow{The last run of this succeeded, and the answer is already in the history. Nothing changed since, and no new question is being asked. Not: repeating after a failure.}
\mbrow{Task}{You are a storage administrator managing disk space for a legacy logging application. ...}
\mbrow{Prefix}{\mbcode{\$ find / -{}type d -{}name \textquotedbl{}localssd\textquotedbl{} 2\textgreater{}/dev/null}\newline\mbout{/mnt/localssd}}
\mbrow{Step}{\mbthink{Good. Let me check the system configuration more carefully.}\newline\mberr{\$ find / -{}type d -{}name \textquotedbl{}localssd\textquotedbl{} 2\textgreater{}/dev/null}}
\mbrowlast{Localizer}{The agent already listed /mnt/ and confirmed /mnt/localssd exists, so this find adds no new information and nothing changed in between.}
}%
\mbexample{Retries without fixing the cause}{11.2\%}{%
\mbdefrow{The last run failed, and the agent runs it again without removing the cause. Cosmetic edits, a shorter timeout, or extra logging do not count as a change. Not: repeating a successful check.}
\mbrow{Task}{You are a forensics analyst investigating a compromised Linux host. ...}
\mbrow{Prefix}{\mbcode{\$ /app/jshon-{}20131105/jshon -{}e cert\_b64 /home/user/evidence/1.json}\newline\mbout{warning: nothing to read\newline internal error: stack underflow\newline (exit\_code=1)}}
\mbrow{Step}{\mbthink{Let me extract differently.}\newline\mberr{\$ /app/jshon-{}20131105/jshon -{}e cert\_b64 /home/user/evidence/1.json 2\textgreater{}\&1}}
\mbrowlast{Localizer}{The agent is re-{}running the exact same command that just failed, only adding 2\textgreater{}\&1 which doesn\textquotesingle{}t change the command\textquotesingle{}s behavior or output content since both streams were already visible in the previous run.}
}%
\mbexample{Fails to produce runnable commands}{5.1\%}{%
\mbdefrow{The step contains no executable command. It is prose, an unfinished reasoning block, or tool-{}protocol markup where the command should be. The failure happens before the shell is reached. Not: a command that reaches the shell and fails to parse. Not: a command that runs and prints instead of acting.}
\mbrow{Task}{You are an MLOps engineer tasked with building a tool to track and summarize cross-{}validation experiments ...}
\mbrow{Prefix}{\mbcode{\$ cat \textless{}\textless{}\textquotesingle{}EOF\textquotesingle{} \textbar{} python3 /home/user/tracker.py ...}\newline\mbout{[\{\textquotedbl{}run\_id\textquotedbl{}: \textquotedbl{}auna\textquotedbl{}, \textquotedbl{}mean\textquotedbl{}: 1.0, \textquotedbl{}variance\textquotedbl{}: 0.0\}, ...]}}
\mbrow{Step}{\mbthink{The tie-{}breaking is working correctly -{} \textquotedbl{}auna\textquotedbl{} comes before \textquotedbl{}brun\textquotedbl{} alphabetically. Let me also test for float scores to make sure those work too.}\newline\mberr{\$ (no valid command)}}
\mbrowlast{Localizer}{The step contains no executable command, so it cannot make progress or verify anything.}
}%
\end{tcbraster}
\caption{Examples of misbehavior steps (1/2). Each example shows the definition of the misbehavior, an excerpt of the task, the last step of the prefix, the error step of the policy, and the explanation of the error localizer. The percentage is the share of error steps with the misbehavior. Long text is cut with an ellipsis.}
\label{fig:mb-examples-1}
\end{figure*}

\begin{figure*}[t]
\begin{tcbraster}[raster columns=2, raster equal height=rows, raster column skip=8pt, raster row skip=6pt]
\mbexample{Kills processes without checking their identities}{4.7\%}{%
\mbdefrow{Terminates by a broad name pattern or an unverified process id, without establishing that the matched processes are the intended targets. Not: stopping a process that was identified first.}
\mbrow{Task}{You are a systems programmer debugging a web security WAF (Web Application Firewall) service. ...}
\mbrow{Prefix}{\mbcode{\$ for pid in \$(ls /proc \textbar{} grep -{}E \textquotesingle{}\textasciicircum{}[0-{}9]+\$\textquotesingle{}); do ... done}\newline\mbout{(no output)}}
\mbrow{Step}{\mbthink{Let me try a different approach -{} changing to port 5001 and see if Flask works, or start natively again.}\newline\mberr{\$ pkill -{}9 redis-{}server; pkill -{}9 python3; sleep 2}}
\mbrowlast{Localizer}{The agent\textquotesingle{}s stated plan is to switch Flask to port 5001, yet the command kills redis-{}server and python3 instead of the unknown process holding port 5000, so the port conflict will persist and redis will be down unnecessarily.}
}%
\mbexample{Treats guesses as facts}{2.5\%}{%
\mbdefrow{Commits a guessed or badly extracted value into the work as if it were known, when the history shows the evidence for it is missing, garbled, or contradictory. Not: a hypothesis that is then tested. Not: a constant the task specified.}
\mbrow{Task}{I am a researcher organizing a dataset of academic citations. ... I have a screenshot of the missing ...}
\mbrow{Prefix}{\mbcode{\$ sqlite3 /home/user/datasets/research\_graph.db \textquotedbl{}SELECT source\_id, target\_id FROM edges;\textquotedbl{}}\newline\mbout{1\textbar{}2  5\textbar{}6  1\textbar{}7  7\textbar{}8  2\textbar{}3}}
\mbrow{Step}{\mbthink{Good, edge (2,3) is now in the database. Now I need to add (3,4) and (4,5): P\_2050 (node 3) -{}\textgreater{} P\_3333 (node 4), P\_3333 (node 4) -{}\textgreater{} P\_4004 (node 5).}\newline\mberr{\$ sqlite3 /home/user/datasets/research\_graph.db \textquotedbl{}INSERT INTO edges (source\_id, target\_id) VALUES (3, 4);\textquotedbl{}}}
\mbrowlast{Localizer}{The agent inserted edge (3,4) based on OCR interpretation without verifying the ambiguous readings (2080 vs 2050, 3933 vs 3333) despite acknowledging the uncertainty, which risks inserting an incorrect edge into the database.}
}%
\mbexample{Claims success without verification}{1.7\%}{%
\mbdefrow{The agent declares the task done, or a step successful, without running the check that would show it, or after a check that failed. Not: running a check that misses its target. Not: committing a guessed value.}
\mbrow{Task}{You are a backup operator automating restore tests. ...}
\mbrow{Prefix}{\mbcode{\$ (no command)}\newline\mbout{ERROR: no valid bash tool call in your reply.}}
\mbrow{Step}{\mbthink{I got an error about a missing bash tool call. Let me submit the task now.}\newline\mberr{\$ echo COMPLETE\_TASK\_AND\_SUBMIT\_FINAL\_OUTPUT}}
\mbrowlast{Localizer}{The agent prematurely submitted the task while the daemon is not running (port 8888 is held by an unknown listener, the daemon never successfully bound), so the restore pipeline is unverified and likely broken.}
}%
\mbexample{Prints a binary file as text}{1.4\%}{%
\mbdefrow{Sends the raw bytes of an executable, library, image, or archive to a text display in order to inspect it. Not: a hex dump or an extracted string, which are deliberate.}
\mbrow{Task}{You are a network security engineer investigating a series of suspicious payloads captured on your internal network. ...}
\mbrow{Prefix}{\mbcode{\$ ls -{}la /opt/certs/}\newline\mbout{... decoy\_root.pem\newline ... true\_root.pem}}
\mbrow{Step}{\mbthink{Good, there are two certificates: decoy\_root.pem and true\_root.pem. Let me look at the policy PNG format and extract the fingerprint from it.}\newline\mberr{\$ cat /app/policy.png}}
\mbrowlast{Localizer}{Cats a binary PNG image directly to the terminal, which will dump unreadable binary data and is not a useful way to inspect an image.}
}%
\end{tcbraster}
\caption{Examples of misbehavior steps (2/2). Each example shows the definition of the misbehavior, an excerpt of the task, the last step of the prefix, the error step of the policy, and the explanation of the error localizer. The percentage is the share of error steps with the misbehavior. Long text is cut with an ellipsis.}
\label{fig:mb-examples-2}
\end{figure*}

\subsection{Statistics}

Error profiling rolls out the policy twice on each of the 2,017 environments and yields 12,746 error steps.
Table~\ref{tab:steps} gives, for each misbehavior, its share of these error steps and of the environments in which it appears.
The three most frequent misbehaviors each cover more than 11\% of the error steps and appear in more than a third of the environments.
Even the least frequent one, ``Prints a binary file as text'', appears in 140 environments (6.9\%).
A step can carry more than one label, so the eight misbehaviors together cover 52.1\% of the error steps, slightly less than the sum of their shares.

\begin{table}[H]
\centering
\small
\setlength{\tabcolsep}{4pt}
\begin{tabularx}{\columnwidth}{@{}rXrr@{}}
\toprule
\# & \textbf{Misbehavior} & \textbf{Steps} & \textbf{Envs.} \\
\midrule
1 & Makes false claims about tests & 15.4\% & 46.7\% \\
2 & Repeats unnecessary checks & 11.5\% & 42.2\% \\
3 & Retries without fixing the cause & 11.2\% & 36.5\% \\
4 & Fails to produce runnable commands & 5.1\% & 19.4\% \\
5 & Kills processes without checking their identities & 4.7\% & 14.0\% \\
6 & Treats guesses as facts & 2.5\% & 11.0\% \\
7 & Claims success without verification & 1.7\% & 8.9\% \\
8 & Prints a binary file as text & 1.4\% & 6.9\% \\
\midrule
 & Eight together & 52.1\% & 75.4\% \\
\bottomrule
\end{tabularx}
\caption{Share of the 12,746 error steps (Steps) and of the 2,017 profiled environments (Envs.) for each misbehavior. A step or an environment can have more than one misbehavior.}
\label{tab:steps}
\label{tab:tasks}
\end{table}

\subsection{Examples of Misbehaviors}

Figures~\ref{fig:mb-examples-1} and~\ref{fig:mb-examples-2} show one error step for each misbehavior.
We chose each example for a short prefix and command, so that the error is visible without the full trajectory.
Each box gives the definition written by the annotator, the start of the task, the last step of the prefix, and the error step, whose command is shown in red.
The last row gives the explanation of the error localizer.

\subsection{Long-Tail Error Steps}

The remaining error steps have no label or fall into rare types, each covering less than 1\% of the error steps.
Table~\ref{tab:tail} lists eight examples, together with the detail of the environment that each one depends on.
Fixing such an error requires knowing that detail, such as a log format or a version rule, so the fix is unlikely to help on other environments.

\begin{table}[H]
\centering
\small
\renewcommand{\arraystretch}{1.1}
\setlength{\tabcolsep}{4pt}
\begin{tabularx}{\columnwidth}{@{}r>{\raggedright\arraybackslash}X>{\raggedright\arraybackslash}p{0.3\columnwidth}@{}}
\toprule
\# & \textbf{Long-tail error step} & \textbf{Environment detail} \\
\midrule
1 & Uses a greedy regex that treats an entire log line as one key-value pair. & the format of the audit log \\
2 & Starts frame counting at 1 instead of 0, shifting every 10-frame bucket. & the bucket rule \\
3 & Replaces udb inside existing udbuf, producing udbufbuf. & the variable names in the C file \\
4 & Extracts metric names instead of values with an incorrect awk field index. & the \texttt{NAME=...VALUE=...} line format \\
5 & Reads little-endian WAV samples as big-endian integers. & the audio file \\
6 & Rejects version 1.5.0 despite the \textasciicircum 1.4.2 requirement allowing it. & the version rule \\
7 & Divides the feature sum by 3 instead of the number of valid features. & the imputation rule \\
8 & Ignores the number of decimal places and always divides by 100 rather than 10, turning 25.5 into 2.55. & the number format \\
\bottomrule
\end{tabularx}
\caption{Examples of long-tail error steps. Each comes from a detail of one environment and rarely carries over to other environments.}
\label{tab:tail}
\end{table}

\section{Cost Calculation Details}
\label{app:cost}
\label{app:efficiency}

We count the training resources of each recipe up to its best checkpoint rather than over its full training run, which is fairer to recipes that train longer than needed; for example, the released TMax-4B checkpoint is from step 200 of 500, as stated on its model card.
We also count the resources of all preprocessing stages, such as error profiling for MiRL, step profiling for PivotRL, and rubric writing for RaR+.
Most values are measured from our logs; below we explain the values that are not.

\paragraph{TMax-4B.}
The TMax paper reports the training settings and curves for TMax-4B, but not the total number of rollouts, tokens, or GPU hours.
We estimate these totals from the reported settings and curves.
The number of rollouts is $S B G / (1-d)$, where $S = 200$ steps, $B = 8$ environments per step, $G = 32$ rollouts per environment, and $d$ is the share of dead groups, which are resampled.
We derive $d = 32.5\%$ from Figure~11 in the TMax paper~\citep{ivison2026tmax}, which counts the groups in which all rollouts fail; TMax notes that groups in which all rollouts pass are rare and does not plot them, so we count them as zero.
This gives 75.9K rollouts and $S B / (1-d) = 2{,}370$ environments.
We estimate its tokens and GPU hours from the tokens per step, steps per rollout, and throughput of our TMax-189 run, except for the output tokens per step, which come from Figure~6 in the TMax paper.\looseness=-1

\paragraph{Total cost.}
We convert the tokens of the policy into dollars at the Tinker prices in Table~\ref{tab:prices}, taking the uncached share of the input tokens as two divided by the average number of steps per rollout, and add the model API fee.
This prices the policy as a service billed per token rather than as a rented GPU cluster, so we report the GPU hours separately and do not add them to the total cost.

\begin{table}[tb]
\centering
\small
\setlength{\tabcolsep}{4pt}
\begin{tabular}{@{}lrrr@{}}
\toprule
\textbf{Model} & \textbf{Input} & \textbf{Cached} & \textbf{Output} \\
\midrule
Qwen3.5-4B & \$0.33 & \$0.066 & \$1.005 \\
Qwen3.8-27B & \textyen 3.646 & \textyen 0.729 & \textyen 21.875 \\
Gemini 3.8 Flash & \$0.75 & \$0.075 & \$3.75 \\
Claude Opus 5 & \$5 & \$0.50 & \$25 \\
\bottomrule
\end{tabular}
\caption{Prices per million tokens at the time of our experiments: Qwen3.5-4B (policy) on Tinker, Qwen3.8-27B (error localizer and teacher) on Alibaba Cloud, Gemini 3.8 Flash (judges) from Google, and Claude Opus 5 (annotator) from Anthropic.}
\label{tab:prices}
\end{table}

\section{Artifacts, Licenses, and Use of AI Assistants}
\label{app:artifacts}

\paragraph{Licenses.}
We use the following artifacts under their licenses: the Qwen3.5-4B and Qwen3.8-27B models (Apache 2.0); the TMax environments (ODC-BY) and the TMax-4B checkpoint (Apache 2.0); Terminal-Bench and OpenThoughts-TBLite (Apache 2.0); the open-instruct training code (Apache 2.0) and the mini-SWE-agent harness (MIT) used by TMax; and the judge prompt of NeMo-Gym (Apache 2.0).
We access Gemini 3.8 Flash and Claude Opus 5 through their commercial APIs under the providers' terms of service.
These licenses allow research use, and we use each artifact for its intended research purpose.
We release our code, checkpoint, data, and evaluation trajectories under licenses compatible with the artifacts they build on: Apache 2.0 for code, the checkpoint, and the trajectories, and ODC-BY for data derived from the TMax environments. See \url{https://github.com/runchu-tian/MiRL}.

\paragraph{Packages.}
Table~\ref{tab:packages} lists the software that we use and its settings; Table~\ref{tab:hparams} gives the remaining hyperparameters.

\begin{table}[H]
\centering
\small
\setlength{\tabcolsep}{4pt}
\begin{tabular}{@{}lll@{}}
\toprule
\textbf{Package} & \textbf{Version} & \textbf{Use} \\
\midrule
open-instruct & \texttt{7387d2f9} & RL training (\texttt{grpo\_fast}) \\
DeepSpeed & 0.18.4 & RL training \\
Ray & 2.53.0 & RL training \\
PyTorch & 2.10.0 & RL training \\
Transformers & 5.4.0 & RL training \\
vLLM & 0.19.1 & Inference (bf16) \\
Harbor & 0.6.6 & Evaluation \\
Apptainer & 1.3.4 & Sandbox \\
LiteLLM & 1.75.0 / 1.83.14 & LLM calls (train / eval) \\
difflib & stdlib & PivotRL step matching \\
\midrule
\textbf{Model / prompt} & \textbf{Revision} & \\
\midrule
Qwen3.5-4B & \texttt{851bf6e8} & Policy \\
Qwen3.8-27B & \texttt{1d4bf0f2} & Localizer, teacher, judge \\
NeMo-Gym & \texttt{325fb49d} & PivotRL judge prompt \\
\bottomrule
\end{tabular}
\caption{Software and model versions. open-instruct and the TMax agent come from the same TMax commit. vLLM uses prefix caching and the \texttt{qwen3\_xml} tool-call parser; Harbor allows at most 64 format errors; PivotRL matches steps with \texttt{SequenceMatcher} at a threshold of 0.95; Qwen3.8-27B as the misbehavior-rate judge uses thinking with low effort.}
\label{tab:packages}
\end{table}

\paragraph{Use of AI assistants.}
We used Claude Code, a coding agent, in the terminal to help write Python and shell scripts and to format the LaTeX source.

\end{document}